\documentclass{article}

\usepackage[english]{babel}
\usepackage[utf8]{inputenc} % Use utf8 input encoding
\usepackage[T1]{fontenc} % Use T1 font encoding

\usepackage[letterpaper,top=2cm,bottom=2cm,left=3cm,right=3cm,marginparwidth=1.75cm]{geometry}

\usepackage{amsmath}
\usepackage{amssymb} % For mathematical symbols like \CIRCLE
\usepackage{amsthm} % For theorem environments
\newtheorem{theorem}{Theorem}
\usepackage{graphicx}
\usepackage{url}
\usepackage{pifont} % For dingbats in table

\usepackage[table]{xcolor}

\usepackage{tikz} % For diagrams
\usetikzlibrary{arrows.meta, positioning, shapes, fit, backgrounds, calc, matrix} % TikZ libraries used in diagrams.tex
\usepackage{pgfplots} % For bar charts and plots
\pgfplotsset{compat=1.18} % Set compatibility level
\usepackage{pdflscape} % For landscape orientation

\usepackage{listings} % For code listings
\usepackage{booktabs}    % For professional-looking rules (\toprule, \midrule, \bottomrule)
\usepackage{tabularx}    % For flexible width tables (X column)
\usepackage{ragged2e}    % For \RaggedRight or \Centering in column definitions
\usepackage{caption}     % For customizing captions
\usepackage{multirow}    % For table cells spanning multiple rows

\definecolor{cMany}{HTML}{7B241C}   % Lighter red
\definecolor{cNative}{HTML}{C0392B} % Darker/maroon red

\tikzset{
    base node/.style={rectangle, rounded corners, draw=black, minimum width=2.7cm, minimum height=1.2cm, align=center, font=\sffamily\bfseries},
    data node/.style={base node, fill=orange!20},
    compute node/.style={base node, fill=blue!20},
    flow node/.style={base node, fill=green!20, minimum width=2.2cm}, % Slightly narrower for flow labels
    label style/.style={font=\sffamily\small\itshape, midway},
    arrow style/.style={thick, -{Stealth[length=3mm, width=2mm]}}, % Standardized arrow head
    dashed arrow style/.style={arrow style, dashed, draw=black!70},
    bg box/.style={draw=black!50, dashed, rounded corners, inner sep=0.8cm}, % Consistent background box style
    node distance/.style={node distance=1.5cm and 2cm} % Consistent spacing for positioning library
}

\newcommand{\nodeLifecycleDiagram}{
\begin{tikzpicture}
    \node[draw=black!80, fill=white, rounded corners, text width=2.8cm, align=center, thick] (shared1) {\textbf{Shared Store}\\{\footnotesize Input data}};
    
    \node[draw=black!80, fill=blue!15, rounded corners, text width=2.8cm, align=center, below=1.8cm of shared1, thick] (prep) {\textbf{prep()}\\{\footnotesize Read inputs}};
    \node[draw=black!80, fill=blue!20, rounded corners, text width=2.8cm, align=center, right=2.2cm of prep, thick] (exec) {\textbf{exec()}\\{\footnotesize Process data}};
    \node[draw=black!80, fill=blue!15, rounded corners, text width=2.8cm, align=center, right=2.2cm of exec, thick] (post) {\textbf{post()}\\{\footnotesize Return actions}};
    
    \node[draw=black!80, fill=white, rounded corners, text width=2.8cm, align=center, below=1.8cm of post, thick] (shared2) {\textbf{Shared Store}\\{\footnotesize Output data}};
    
    \draw[rounded corners=5pt, dashed, draw=black!60, line width=1pt] 
          ($(prep.north west)+(-0.5,0.3)$) rectangle ($(post.south east)+(0.5,-0.3)$);
    
    \node[font=\small\itshape, text=black!70] at ($(post.north east)+(-1.5,0.1)$) {Node boundary};
    
    \draw[-stealth, thick, blue!70] (shared1) -- node[font=\small\itshape, text=blue!70, right, pos=0.4] {Read} (prep);
    \draw[-stealth, thick, blue!70] (prep) -- node[font=\small\itshape, text=blue!70, above] {Data} (exec);
    \draw[-stealth, thick, blue!70] (exec) -- node[font=\small\itshape, text=blue!70, above] {Result} (post);
    \draw[-stealth, thick, blue!70] (post) -- node[font=\small\itshape, text=blue!70, right, pos=0.4] {Write} (shared2);
    
    \draw[-stealth, thick, dashed, red!70] 
          (post.south) -- ++(0,-0.6) -- 
          node[font=\small\itshape, text=red!70, below] {Next node} 
          ++(-7.2,0) -- (prep.south);
          
    \node[above=0.5cm of shared1, font=\sffamily\bfseries\large] {Node Lifecycle};
\end{tikzpicture}
}

\newcommand{\flowNestingDiagram}{
\begin{tikzpicture}[scale=0.9, transform shape] % Scale down to prevent overflow
    \tikzset{
        flow/.style={
            draw=black!80, 
            fill=green!15, 
            rounded corners=3pt, 
            text width=2.2cm, % Reduced width
            align=center, 
            minimum height=0.9cm, % Reduced height
            font=\sffamily\bfseries
        },
        node/.style={
            draw=black!80, 
            fill=blue!15, 
            rounded corners=3pt, 
            text width=2.2cm, % Reduced width
            align=center, 
            minimum height=0.9cm, % Reduced height
            font=\sffamily
        },
        arrow/.style={
            -stealth, 
            thick,
            black!70
        },
        dashed arrow/.style={
            -stealth, 
            thick,
            dashed,
            black!70
        },
        num label/.style={
            font=\small\itshape,
            fill=white,
            inner sep=1pt,
            text=black!80
        }
    }
    
    \node[flow] (parent) at (0,0) {Parent Flow};
    \node[node, below left=1.6cm and 1cm of parent] (nodeA) {Node A};
    \node[flow, below=2.5cm of parent] (subflow) {Sub-Flow};
    \node[node, below right=1.6cm and 1cm of parent] (nodeC) {Node C};
    
    \draw[arrow] (parent) to[out=-150, in=60] node[num label, pos=0.4] {1} (nodeA);
    \draw[arrow] (nodeA) to[out=-30, in=150] node[num label, pos=0.5] {2} (subflow);
    \draw[arrow] (subflow) to[out=30, in=-150] node[num label, pos=0.5] {3} (nodeC);
    \draw[arrow] (nodeC) to[out=60, in=-30, looseness=1.2] node[num label, pos=0.6] {4} (parent);
    
    \begin{scope}[xshift=7cm] % Reduced spacing between diagrams
        \node[flow] (sub) at (0,0) {Sub-Flow};
        \node[node, below left=1.6cm and 1cm of sub] (nodeB1) {Node B1};
        \node[node, below right=1.6cm and 1cm of sub] (nodeB2) {Node B2};
        
        \draw[arrow] (sub) to[out=-150, in=60] node[num label, pos=0.4] {1} (nodeB1);
        \draw[arrow] (nodeB1) to[out=0, in=180] node[num label, pos=0.5] {2} (nodeB2);
        \draw[arrow] (nodeB2) to[out=60, in=-30, looseness=1.2] node[num label, pos=0.6] {3} (sub);
    \end{scope}
    
    \draw[dashed arrow] (subflow) -- node[font=\small\itshape, fill=white, inner sep=2pt] {Expands to} (sub);
    
    \begin{pgfonlayer}{background}
        \filldraw[rounded corners=5pt, fill=green!5, draw=green!40, dashed] 
            ($(parent.north west)+(-0.5,0.3)$) rectangle ($(nodeC.south east)+(0.5,-0.3)$);
            
        \filldraw[rounded corners=5pt, fill=green!5, draw=green!40, dashed] 
            ($(sub.north west)+(-0.5,0.3)$) rectangle ($(nodeB2.south east)+(0.5,-0.3)$);
    \end{pgfonlayer}
    
    \node[font=\sffamily\bfseries, above=0.3cm of parent] {Parent Flow};
    \node[font=\sffamily\bfseries, above=0.3cm of sub] {Sub-Flow (Expanded)};
    
    \node[font=\small\itshape, text width=3.5cm, align=center, anchor=north] 
        at ($(nodeC.south east)!0.5!(nodeB1.south west)+(0,-0.8)$) 
        {Sub-flows are treated as single nodes within parent flows};
\end{tikzpicture}
}

\newcommand{\conditionalTransitionsDiagram}{
\begin{tikzpicture}[scale=0.9, transform shape] % Scale down to prevent overflow
    \tikzset{
        review/.style={
            draw=black!80, 
            fill=yellow!15, 
            rounded corners=3pt, 
            text width=2.5cm, 
            align=center, 
            minimum height=1cm,
            font=\sffamily\bfseries
        },
        action/.style={
            draw=black!80, 
            fill=blue!15, 
            rounded corners=3pt, 
            text width=2.5cm, 
            align=center, 
            minimum height=1cm,
            font=\sffamily
        },
        arrow/.style={
            -stealth, 
            thick,
            black!70
        },
        action label/.style={
            font=\small\itshape,
            fill=white,
            inner sep=1pt,
            text=black!80
        }
    }
    
    \node[review] (review) at (0,0) {Review Node};
    
    \node[action, below left=2cm and 1.5cm of review] (approve) {Process Payment};
    \node[action, below=2.5cm of review] (revise) {Revise Node};
    \node[action, below right=2cm and 1.5cm of review] (reject) {Reject Node};
    
    \draw[arrow] (review) to[out=-150, in=60] 
        node[action label, pos=0.5] {\texttt{"approved"}} (approve);
    \draw[arrow] (review) to[out=-90, in=90] 
        node[action label, pos=0.5] {\texttt{"needs\_revision"}} (revise);
    \draw[arrow] (review) to[out=-30, in=120] 
        node[action label, pos=0.5] {\texttt{"rejected"}} (reject);
    
    \draw[arrow, dashed] (approve) -- ++(0,-1) -| 
        node[action label, pos=0.75, below] {Next flow step} ++(3,0);
    \draw[arrow, dashed] (revise) -- ++(0,-1) -| 
        node[action label, pos=0.25, below] {Back to review} (review);
    \draw[arrow, dashed] (reject) -- ++(0,-1) -| 
        node[action label, pos=0.75, below] {End flow} ++(-3,0);
    
    \begin{pgfonlayer}{background}
        \filldraw[rounded corners=5pt, fill=yellow!5, draw=yellow!40, dashed] 
            ($(review.north west)+(-0.5,0.3)$) rectangle ($(review.south east)+(0.5,-0.3)$);
    \end{pgfonlayer}
    
    \node[font=\sffamily\bfseries\large, above=0.4cm of review] {Conditional Transitions};
    
    \node[font=\small\itshape, text width=5cm, align=center, below=1.0cm of revise] 
        {Action strings from a node's post method determine the next execution path};
\end{tikzpicture}
}

\newcommand{\ragArchitectureDiagram}{
\begin{tikzpicture}[scale=0.9, transform shape]
    \tikzset{
        flow/.style={
            draw=black!80, 
            fill=teal!15, 
            rounded corners=3pt, 
            text width=2.5cm, 
            align=center, 
            minimum height=0.9cm,
            font=\sffamily\bfseries
        },
        node/.style={
            draw=black!80, 
            fill=blue!15, 
            rounded corners=3pt, 
            text width=2.5cm, 
            align=center, 
            minimum height=0.9cm,
            font=\sffamily
        },
        store/.style={
            draw=black!80, 
            fill=gray!10, 
            rounded corners=3pt, 
            text width=7cm, 
            align=center, 
            minimum height=1.2cm,
            font=\sffamily
        },
        arrow/.style={
            -stealth, 
            thick,
            black!70
        },
        data arrow/.style={
            -stealth, 
            thick,
            dashed,
            blue!70
        }
    }
    
    \node[flow] (offline) at (0,0) {Offline Flow};
    \node[node, below=0.8cm of offline] (chunk) {Chunk Documents};
    \node[node, below=0.8cm of chunk] (embed) {Embed Documents};
    \node[node, below=0.8cm of embed] (index) {Create Index};
    \node[node, below=0.8cm of index] (vector) {Vector Index};
    
    \draw[arrow] (offline) -- node[font=\small\itshape, fill=white, inner sep=1pt, right, pos=0.5] {1} (chunk);
    \draw[arrow] (chunk) -- node[font=\small\itshape, fill=white, inner sep=1pt, right, pos=0.5] {2} (embed);
    \draw[arrow] (embed) -- node[font=\small\itshape, fill=white, inner sep=1pt, right, pos=0.5] {3} (index);
    \draw[arrow] (index) -- node[font=\small\itshape, fill=white, inner sep=1pt, right, pos=0.5] {4} (vector);
    
    \begin{scope}[xshift=6cm]
        \node[flow] (online) at (0,0) {Online Flow};
        \node[node, below=0.8cm of online] (query) {Embed Query};
        \node[node, below=0.8cm of query] (retrieve) {Retrieve Docs};
        \node[node, below=0.8cm of retrieve] (generate) {Generate Answer};
        \node[node, below=0.8cm of generate] (answer) {Final Answer};
        
        \draw[arrow] (online) -- node[font=\small\itshape, fill=white, inner sep=1pt, right, pos=0.4] {1} (query);
        \draw[arrow] (query) -- node[font=\small\itshape, fill=white, inner sep=1pt, right, pos=0.5] {2} (retrieve);
        \draw[arrow] (retrieve) -- node[font=\small\itshape, fill=white, inner sep=1pt, right, pos=0.5] {3} (generate);
        \draw[arrow] (generate) -- node[font=\small\itshape, fill=white, inner sep=1pt, right, pos=0.5] {4} (answer);
    \end{scope}
    
    \draw[data arrow] (vector) to[out=0, in=180] 
        node[font=\small\itshape, fill=white, inner sep=1pt, above] {Index data} (retrieve);
    
    \node[store, below=1.5cm of vector, xshift=3cm] (shared) {
        \textbf{Shared Store}\\
        \small{Document chunks, embeddings, index, query, retrieved docs, answer}
    };
    
    \draw[data arrow, blue!50] (chunk.west) -- ++(-0.5,0) |- node[font=\small\itshape, fill=white, inner sep=1pt, pos=0.4, left] {Chunks} (shared.west);
    \draw[data arrow, blue!50] (embed.west) -- ++(-0.7,0) |- node[font=\small\itshape, fill=white, inner sep=1pt, pos=0.3, left] {Embeddings} (shared.west);
    \draw[data arrow, blue!50] (index.west) -- ++(-0.9,0) |- node[font=\small\itshape, fill=white, inner sep=1pt, pos=0.8, left] {Index} (shared.west);
    
    \draw[data arrow, blue!50] (query.east) -- ++(0.7,0) |- node[font=\small\itshape, fill=white, inner sep=1pt, pos=0.4, right] {Query} (shared.east);
    \draw[data arrow, blue!50] (retrieve.east) -- ++(0.9,0) |- node[font=\small\itshape, fill=white, inner sep=1pt, pos=0.3, right] {Docs} (shared.east);
    \draw[data arrow, blue!50] (answer.east) -- ++(1.1,0) |- node[font=\small\itshape, fill=white, inner sep=1pt, pos=0.8, right] {Answer} (shared.east);
    
    \begin{pgfonlayer}{background}
        \filldraw[rounded corners=5pt, fill=teal!5, draw=teal!40, dashed] 
            ($(offline.north west)+(-0.4,0.3)$) rectangle ($(vector.south east)+(0.4,-0.3)$);
            
        \filldraw[rounded corners=5pt, fill=teal!5, draw=teal!40, dashed] 
            ($(online.north west)+(-0.4,0.3)$) rectangle ($(answer.south east)+(0.4,-0.3)$);
    \end{pgfonlayer}
    
    \node[font=\sffamily\bfseries, above=0.3cm of offline] {Offline Indexing};
    \node[font=\sffamily\bfseries, above=0.3cm of online] {Online Query};
    
    \node[font=\sffamily\bfseries\large, above=1.0cm of offline, xshift=3cm] {RAG Architecture with Pocketflow};
\end{tikzpicture}
}

\title{Flow State: Humans Enabling AI Systems to Program Themselves}

\author{
  Helena Zhang\\
  Stanford University\\
  \texttt{helena@pocketflow.ai}
  \and % Separates first and second author
  Jakobi Haskell\\
  Brown University\\
  \texttt{jakobi@pocketflow.ai}
  \and % Separates second and third author
  Yosi Frost\\
  Pocketflow AI\\
  \texttt{yosi@pocketflow.ai}
}

\date{\today} % Or specify a date: \date{April 2025}

\begin{document}
\maketitle

\begin{abstract}
Compound AI systems, orchestrating multiple AI components and external APIs, are increasingly vital but face challenges in managing complexity, handling ambiguity, and enabling effective development workflows. Existing frameworks often introduce significant overhead, implicit complexity, or restrictive abstractions, hindering maintainability and iterative refinement, especially in Human-AI collaborative settings. We argue that overcoming these hurdles requires a foundational architecture prioritizing structural clarity and explicit control. To this end, we introduce \emph{Pocketflow}, a platform centered on Human-AI co-design, enabled by \emph{Pocketflow}. Pocketflow is a Python framework built upon a deliberately minimal yet synergistic set of core abstractions: modular \texttt{Nodes} with a strict lifecycle, declarative \texttt{Flow} orchestration, native hierarchical nesting (\texttt{Flow}-as-\texttt{Node}), and explicit action-based conditional logic. This unique combination provides a robust, vendor-agnostic foundation with very little code that demonstrably reduces overhead while offering the expressiveness needed for complex patterns like agentic workflows and RAG. Complemented by \emph{Pocket AI}, an assistant leveraging this structure for system design, Pocketflow provides an effective environment for iteratively prototyping, refining, and deploying the adaptable, scalable AI systems demanded by modern enterprises.
\end{abstract}

% Input sections - now flattened to the same directory
\section{Introduction}
\label{sec:introduction}

Enterprises commonly face repetitive yet complex operational tasks, such as data entry, report generation, and invoice processing, which incur significant time and cost overheads. While Large Language Models (LLMs) exhibit remarkable capabilities applicable to automation~\cite{bubeck2023sparks}, individual LLM interactions are often insufficient for reliably handling multi-step, real-world processes. Consequently, the paradigm of \emph{Compound AI Systems} has emerged, integrating multiple LLM calls with external APIs and tools to address complex automation challenges~\cite{zaharia2024shift,santhanam2024alto,chen2023interleaving}. Such systems are increasingly deployed across various industries to streamline operations and enhance decision-making.

However, the effective design of compound AI systems presents considerable challenges. Key design decisions include task decomposition strategy~\cite{khot2022decomposed,wies2023subtask}, component selection and integration, LLM call orchestration, the incorporation points for human oversight or feedback~\cite{wang2024human,liang2023holistic}, and the selection of appropriate external APIs or tools. The quality of these decisions directly impacts the accuracy, reliability, and overall efficacy of the resulting system.

Existing approaches to accelerate AI system design, such as automated system optimizers like DSPy~\cite{khattab2024dspy} or DocETL~\cite{shankar2024validates}, attempt to navigate the design space based on predefined input-output specifications. These methods, however, face significant limitations:
\begin{itemize}
    \item \textbf{Restricted Search Space:} Current optimizers often operate within a constrained set of predefined operators (e.g., prompt engineering, basic chaining in DSPy). They typically lack the capability to construct complex agentic behaviors, devise novel retrieval mechanisms~\cite{edge2024local,lewis2020retrieval,asai2023self}, integrate human-in-the-loop feedback dynamically, or incorporate arbitrary external tools and APIs.
    \item \textbf{Requirement Ambiguity:} Real-world tasks frequently involve inherent ambiguity, making it difficult to specify desired outputs or evaluation criteria definitively \emph{a priori}~\cite{shankar2024validates,kandogan2024blueprint}. Iterative refinement, driven by human evaluation of candidate outputs and domain expertise, is often necessary to disambiguate objectives and establish clear success metrics.
\end{itemize}
Therefore, while full automation holds promise for optimization \emph{after} initial requirements are clarified, human insight remains indispensable during the early stages of design and prototyping for task disambiguation and the creation of grounded, innovative solutions.

We posit that the future of complex AI system development lies in \textbf{Human-AI Co-Design}, an iterative process synergizing human expertise with AI capabilities. This collaborative workflow, outlined below and summarized in Table~\ref{tab:process_steps}, allows humans to guide high-level strategy while AI handles implementation details and optimization:
\begin{enumerate}
    \item \textbf{Requirement Specification:} Human stakeholders define high-level objectives, provide domain-specific insights, and identify essential tools or APIs. AI may assist by structuring discussions or drafting initial requirement documents~\cite{zhou2023understanding,yang2022ai}.
    \item \textbf{High-Level System Design:} AI proposes system architectures, workflow structures, data schemas, and potential component interactions, actively soliciting human feedback for refinement. Human engineers ensure the feasibility and correctness of integrating external systems or APIs~\cite{wu2023autogen,li2024self}.
    \item \textbf{Prototyping and Evaluation:} AI rapidly generates code prototypes based on the refined design. Humans provide representative inputs, evaluate system outputs against implicit or explicit criteria grounded in domain expertise, and offer iterative feedback for improvement~\cite{lee2022coauthor,jiang2023evaluating}.
    \item \textbf{System Optimization:} Once the prototype demonstrates feasibility, humans provide more comprehensive input data and define quantitative evaluation metrics. AI utilizes this feedback to optimize system parameters, prompts, or potentially incorporates techniques from automated optimization frameworks~\cite{khattab2024dspy,chen2023program}. AI may also generate test cases and identify performance bottlenecks.
    \item \textbf{Reliable Deployment \& Maintenance:} AI assists in generating deployment configurations (e.g., CI/CD pipelines), error-handling logic, and monitoring setups. Humans oversee the process, handle complex deployment issues, and approve critical fixes, while AI can perform routine monitoring, log analysis, and propose solutions for common problems~\cite{potts2023llmops,li2024verifiers}.
\end{enumerate}

% Note: Ensure the 'ding' symbols render correctly by including the pifont package in main.tex.
\begin{table}[ht]
    \centering
    \caption{Overview of Human-AI Roles Across Co-Design Stages. Involvement level indicated by symbols ($\bullet$=High, $\circ$=Medium, --=Low; subjective scale).} 
    \label{tab:process_steps} 
    \scriptsize 
    \renewcommand{\arraystretch}{1.3} 
    \begin{tabularx}{\textwidth}{@{}%
        >{\RaggedRight\arraybackslash}p{0.16\textwidth}% Stage Name
        >{\centering\arraybackslash}p{0.10\textwidth}% Human Involvement
        >{\centering\arraybackslash}p{0.10\textwidth}% AI Involvement
        >{\RaggedRight\arraybackslash}X % Details
        @{}}
        
        \toprule
        \textbf{Stage}
        & \textbf{Human Role} 
        & \textbf{AI Role} 
        & \textbf{Details of Collaboration} \\
        \midrule

        \textbf{1. Requirements}
        & $\bullet$ % High
        & --    % Low
        & \textbf{Human:} Defines scope, navigates ambiguity, handles business context, resolves conflicting needs.
          \newline \textbf{AI:} Summarizes discussions, drafts initial requirements. \\
        \midrule

        \textbf{2. Design}
        & $\circ$ % Medium
        & $\circ$ % Medium
        & \textbf{Human:} Provides tool/API docs, validates integration points, gives architectural feedback.
          \newline \textbf{AI:} Proposes workflows, data schemas, component interactions. \\
        \midrule

        \textbf{3. Prototyping}
        & $\circ$ % Medium
        & $\circ$ % Medium
        & \textbf{Human:} Reviews code, refines design choices, provides sample inputs, evaluates outputs based on expertise.
          \newline \textbf{AI:} Generates code, performs refactoring, creates documentation, rapidly builds functional prototypes. \\
        \midrule

        \textbf{4. Testing \& Optimization}
        & $\circ$ % Medium
        & $\circ$ % Medium
        & \textbf{Human:} Performs exploratory testing, defines success criteria, provides optimization targets.
          \newline \textbf{AI:} Generates unit/integration tests, mock data; identifies bottlenecks; suggests or applies optimizations based on feedback. \\
        \midrule

        \textbf{5. Deployment}
        & --    % Low
        & $\bullet$ % High
        & \textbf{Human:} Oversees deployment, handles complex incidents/rollbacks.
          \newline \textbf{AI:} Configures CI/CD, sets up monitoring, performs routine deployment tasks, flags anomalies. \\
        \midrule

        \textbf{6. Maintenance}
        & --    % Low
        & $\bullet$ % High
        & \textbf{Human:} Approves significant fixes, handles complex escalations or architectural drift.
          \newline \textbf{AI:} Monitors performance, analyzes logs, proposes fixes for common issues, performs routine updates. \\

        \bottomrule
    \end{tabularx}
\end{table}

While Human-AI co-design is gaining traction in areas like web application development (e.g., using tools like bolt.new~\cite{boltnew2025} or cursor.ai~\cite{cursorai2025}), its application to compound AI system design is hampered by the lack of suitable foundational frameworks. Standard software frameworks (e.g., React, Vue, Node.js, Express.js) provide robust, well-defined abstractions that benefit both human and AI developers, such as reusable components and clear separation of concerns. In contrast, existing frameworks for AI systems~\cite{wu2023autogen,langchain,langgraph,crewai,pydantic,rasmussen2023wellcomposition} often suffer from several drawbacks: they tend to include numerous vendor-specific wrappers, rely on brittle or overly specific abstractions (e.g., specific chain types, text splitters), embed inflexible hard-coded prompts, and accumulate significant dependency bloat~\cite{baseten2025langchain,wang2024aicompanion}. These limitations often lead practitioners to abandon existing frameworks, resulting in ad-hoc, less maintainable solutions. This highlights a critical gap: the need for a foundational architecture that directly confronts the challenges of abstraction mismatch (where convenient high-level constructs become brittle) and implicit complexity (where hidden behaviors obscure logic and hinder debugging). Effectively managing the intricacies of compound AI systems, handling requirement ambiguity through iteration, and enabling productive Human-AI co-design necessitates a return to first principles: modularity, explicit control, and structural clarity. Pocketflow is conceived as a direct answer to this need. Its structure-first, minimal-core design philosophy is not merely about reducing dependencies, but about providing the essential, orthogonal primitives required to build complex, adaptable systems without succumbing to the pitfalls of premature or overly opinionated abstractions.

We introduce \textbf{Pocketflow}, a platform conceived specifically for this Human-AI co-design paradigm for compound AI systems. Pocketflow is built on the hypothesis that a synergistic combination of fundamental abstractions can provide the necessary balance of expressiveness and manageability. Its core components are:
\begin{enumerate}
    \item \textbf{Pocketflow:} Not just a minimalistic framework, but a foundational architecture based on the synergistic interplay of \emph{nested directed graphs}, a strict \emph{Node lifecycle}, and \emph{explicit, declarative control flow}. Its design directly targets the sources of complexity and opacity in AI system development by prioritizing:
        \begin{itemize}
            \item Very little code and minimal dependencies
        \end{itemize}
    \item \textbf{Pocket AI:} To bridge the gap between AI's syntactic proficiency and the semantic requirements of robust software engineering—addressing the common pitfall where AI assistants generate unmaintainable code by simply adding complexity—we introduce Pocket AI. It's an assistant architected for the Human-AI co-design paradigm, leveraging Pocketflow's structure as essential semantic guardrails. Pocket AI utilizes a specialized multi-agent architecture, dividing labor to ensure quality and context-awareness:
        \begin{itemize}
            \item A \textit{Flow Orchestrator Agent} handles high-level planning, interpreting requirements into Pocketflow's explicit graph structures.
            \item A \textit{Node Generator Agent} focuses on implementation, generating code that adheres strictly to the modular \texttt{Node} lifecycle and framework best practices, promoting testability and separation of concerns.
            \item A \textit{Codebase Navigator Agent} provides crucial context, understanding existing utilities and managing interactions within the project environment.
            \item \textbf{Dynamic Tool Integration:} Utilizes on-demand web search to retrieve current API documentation and usage patterns, avoiding hard-coded dependencies and vendor lock-in.
            \item \textbf{Contextual Customization:} Employs Retrieval-Augmented Generation (RAG) against enterprise knowledge bases to incorporate domain-specific logic and requirements.
        \end{itemize}
\end{enumerate}
Together, Pocketflow provides the robust, minimal foundation, while Pocket AI delivers the adaptive intelligence needed to apply this foundation effectively in complex, real-world scenarios. This structured approach allows Pocket AI to assist in generating solutions that are not just functional but also well-architected and maintainable, directly tackling the limitations observed in less structured AI code generation. The result is a platform that enables a more effective Human-AI co-design process, where humans can focus on high-level architecture and requirements while AI handles implementation details within a framework that enforces good software engineering practices.
% ==================================================
% SECTION 2: Pocketflow Foundational Abstractions
% ==================================================
% This section incorporates the previously revised version.
\section{Pocketflow: Foundational Abstractions}
\label{sec:core_abstractions}

The engineering of robust and maintainable AI systems necessitates carefully designed abstractions that balance expressiveness with manageable complexity~\cite{zaharia2024shift, kandogan2024blueprint, santhanam2024alto}. Pocketflow introduces a minimalistic yet powerful set of foundational abstractions designed to facilitate the construction of complex AI workflows, particularly within a Human-AI co-design paradigm~\cite{wang2024human, yang2022human, zhou2023understanding}. This section details these core components: Nodes, Flows, the native nesting mechanism, and the explicit conditional logic system.

\subsection{Nodes: Encapsulated Units of Computation}

The fundamental building block in Pocketflow is the \texttt{Node} (typically inheriting from \texttt{BaseNode}). Conceptually, a \texttt{Node} represents an atomic, self-contained unit of computation or logic within a larger directed acyclic graph (DAG) representing the workflow. This design is grounded in principles of modularity, promoting separation of concerns as advocated by Parnas~\cite{parnas1972criteria} and aligning with component-based software engineering practices~\cite{khot2022decomposed,weller2022component,lu2021codegen}.

Each \texttt{Node} adheres to a strict three-phase execution lifecycle, enforcing a clear distinction between data handling and core computation:

\begin{enumerate}
    \item \textbf{\texttt{prep(shared)}:} This phase focuses solely on input acquisition. It reads necessary data dependencies from the \texttt{shared} data store (typically a Python dictionary functioning as shared memory across the workflow) and prepares the inputs required for the node's primary logic. Isolating data retrieval simplifies dependency management and enhances testability~\cite{palacio2023contextual,wei2022chainofthought}.
    \item \textbf{\texttt{exec(prep\_res)}:} This phase performs the node's core computational task using only the prepared inputs from the \texttt{prep} phase. It has no direct access to the \texttt{shared} store, ensuring functional purity within this step. This is where interactions with external services (e.g., LLM APIs, databases, tools) occur, often encapsulated within utility functions. The \texttt{BaseNode} implementation incorporates optional, configurable retry logic (\texttt{max\_retries}, \texttt{wait}) and a fallback mechanism (\texttt{exec\_fallback}) to enhance operational robustness without cluttering the core logic~\cite{potts2023llmops,ribeiro2023adaptive}.
    \item \textbf{\texttt{post(shared, prep\_res, exec\_res)}:} Following successful execution, this phase integrates the results back into the system state. It writes outputs to the \texttt{shared} data store and determines the subsequent control flow by returning a string-based \texttt{action} code (defaulting to \texttt{"default"})~\cite{hallacy2023ai,chen2023interleaving}.
\end{enumerate}

This clear separation not only aids traditional debugging but also provides well-defined boundaries for AI code generation tools. In a co-design context, humans can focus on the high-level logic in the \texttt{exec} phase while AI assistants can handle the standardized \texttt{prep} and \texttt{post} phases, or vice versa, creating natural collaboration points~\cite{lee2022coauthor,jiang2023evaluating,zhou2023understanding}.

Critically, the \texttt{action} string returned by the \texttt{post} method serves as the explicit signal for directing workflow execution. This mechanism forms the basis of Pocketflow's declarative conditional logic, detailed further in Section~\ref{ssec:conditional_logic}.

A canonical example illustrates this lifecycle (Listing~\ref{lst:greet_node}):
\begin{lstlisting}[language=Python, caption={Minimal Node Example illustrating the lifecycle.}, label={lst:greet_node}]
from pocketflow import Node # Assuming pocketflow library structure

class GreetNode(Node):
    """A simple node demonstrating the prep-exec-post lifecycle."""

    def prep(self, shared: dict) -> str:
        """Reads 'name' from shared state, provides default."""
        # Input: Shared state dictionary
        # Output: Name string for exec phase
        return shared.get("name", "World")

    def exec(self, name: str) -> str:
        """Executes the core logic: creating a greeting."""
        # Input: Name string from prep_res
        # Output: Greeting string for post phase
        greeting = f"Hello, {name}!"
        # Consider structured logging instead of print in production
        print(f"Executing GreetNode: Generated '{greeting}'")
        return greeting

    def post(self, shared: dict, prep_res: str, exec_res: str) -> str:
        """Writes the greeting back to shared state."""
        # Input: Shared state, prep result (name), exec result (greeting)
        # Output: Action string ("default" for sequential flow)
        shared["greeting"] = exec_res
        return "default" # Proceed to the default next node
\end{lstlisting}
This disciplined separation of data access, computation, and state updates within each \texttt{Node} enhances modularity, facilitates unit testing, simplifies debugging, and makes components more comprehensible for both human developers and AI code generation/analysis tools.

\subsection{Flows: Declarative Workflow Orchestration}

A \texttt{Flow} object orchestrates the execution of an interconnected graph of \texttt{Nodes}. It embodies the workflow structure itself, separating the high-level process definition from the implementation details encapsulated within individual \texttt{Nodes}. This aligns with established practices in workflow management systems and DAG-based task schedulers~\cite{bubeck2023sparks, khattab2024dspy, subbarao2024largelanguage, qian2023communicative}, promoting clarity compared to approaches where control flow logic might be embedded within computational components.

The \texttt{Flow} manages execution based on the \texttt{action} strings returned by node \texttt{post} methods:
\begin{itemize}
    \item A \texttt{Flow} is initialized with a starting \texttt{Node}.
    \item It propagates a \texttt{shared} data store (typically a dictionary) throughout the execution, enabling communication and state persistence between \texttt{Nodes}.
    \item Upon completion of a node's \texttt{post} method, the \texttt{Flow} uses the returned \texttt{action} string to determine the next \texttt{Node} by consulting the current node's \texttt{successors} mapping (an internal dictionary linking action strings to successor nodes).
    \item Execution proceeds through the graph until a node returns an \texttt{action} that does not map to a successor in its \texttt{successors} dictionary, or until a terminal node (one with no defined successors) is reached.
\end{itemize}
The connection topology is defined declaratively using Python operators: \texttt{>>} signifies the default transition (linked to the \texttt{"default"} action), while the \texttt{node - "action\_name" >> next\_node} syntax establishes conditional transitions linked to specific action strings~\cite{wu2023autogen, li2024self, marasovic2022fewshot}.

This declarative approach provides significant advantages for Human-AI co-design. The explicit graph structure serves as a shared representation that humans can easily sketch or visualize, while AI assistants can programmatically parse and generate. This creates a natural division of labor where humans can focus on high-level workflow design while AI implements the individual node logic, or where AI suggests workflow optimizations that humans can readily understand and evaluate~\cite{yang2022human, zhang2024collagenqa, xu2023expertprompting, zhou2023understanding}.

Consider a simple sequence (Listing~\ref{lst:seq_flow}):
\begin{lstlisting}[language=Python, caption={Defining and executing a sequential Flow.}, label={lst:seq_flow}]
from pocketflow import Flow
# Assume GreetNode and AskMoodNode classes are defined as above

# Instantiate nodes
greet_node = GreetNode()
ask_mood_node = AskMoodNode() # Assumed defined elsewhere

# Define the workflow topology: greet_node transitions to ask_mood_node by default
greet_node >> ask_mood_node

# Create the Flow instance, specifying the entry point
greeting_flow = Flow(start=greet_node)

# Execute the flow, providing initial shared state
shared_data = {"name": "Alice"}
final_state = greeting_flow.run(shared=shared_data) # run returns the final state

print(f"Final shared data: {final_state}")
# Example output: {'name': 'Alice', 'greeting': 'Hello, Alice!', 'mood': 'Happy'}
# (Actual output depends on AskMoodNode's execution)
\end{lstlisting}
This declarative approach clearly separates the definition of the workflow graph from the node implementations, enhancing readability and maintainability.

\subsection{Native Nesting: Hierarchical Workflow Composition}

A cornerstone of Pocketflow's design for managing complexity is its mechanism for \textbf{native hierarchical composition}. Because the \texttt{Flow} class itself inherits from the \texttt{BaseNode} interface, an entire \texttt{Flow} can be treated as a single computational unit (\texttt{Node}) within a higher-level \texttt{Flow}. This allows complex sub-workflows to be encapsulated and reused seamlessly.

\begin{figure}[ht]
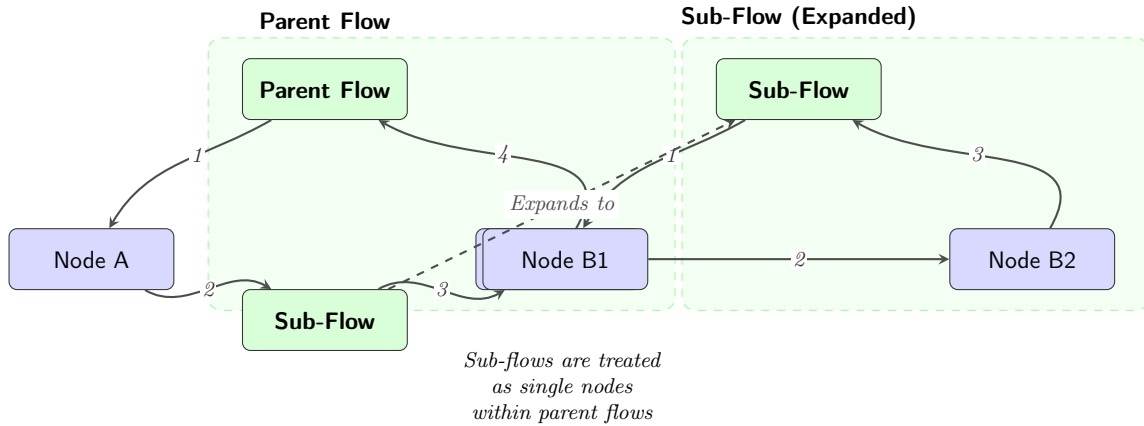

    \centering
    \flowNestingDiagram
    \caption{Flow Nesting: Illustration of hierarchical composition where a Sub-Flow is treated as a single Node within a Parent Flow.}
    \label{fig:flow_nesting}
\end{figure}

This design contribution draws inspiration from hierarchical state machines and modular system architectures~\cite{bubeck2023sparks, shankar2024validates, chen2023program, rasmussen2023wellcomposition} but applies the concept directly and uniformly to workflow orchestration. It provides an integrated mechanism for abstraction and reuse, potentially overcoming the boilerplate or conceptual disconnect associated with ad-hoc sub-graph definitions in other frameworks~\cite{kandogan2024blueprint, wang2024aicompanion, lewis2020retrieval}. The ability to abstract a complex process into a single, reusable component significantly aids in managing the cognitive load associated with large-scale system design (Figure~\ref{fig:flow_nesting}).

This uniform hierarchical composition is particularly valuable for Human-AI co-design. It allows humans to focus on different levels of abstraction as needed—either the high-level system architecture or specific sub-flow implementations. For AI assistants, it simplifies code generation by providing clear compositional boundaries and reuse patterns~\cite{liang2023holistic, asai2023self, wies2023subtask, zhou2023understanding}.

Example of composing flows (Listing~\ref{lst:nested_flow}):
\begin{lstlisting}[language=Python, caption={Nesting a Flow within another Flow.}, label={lst:nested_flow}]
# Assume process_data_flow is a previously defined Flow instance
# Assume analyze_results_node is a Node instance

# process_data_flow can be treated like any other node
# Define the higher-level pipeline topology
process_data_flow >> analyze_results_node

# Create the composite Flow
# The entire process_data_flow acts as the starting "node"
data_pipeline = Flow(start=process_data_flow)

# Execute the composite pipeline
initial_data = {"raw_data": [...]}
final_state = data_pipeline.run(shared=initial_data)
print(f"Pipeline finished. Final state: {final_state}")
\end{lstlisting}
This native nesting capability fosters a divide-and-conquer approach, enabling modular development and simplifying the overall system architecture.

\subsection{Explicit Conditional Logic via Action Strings}
\label{ssec:conditional_logic}

\begin{figure}[ht]
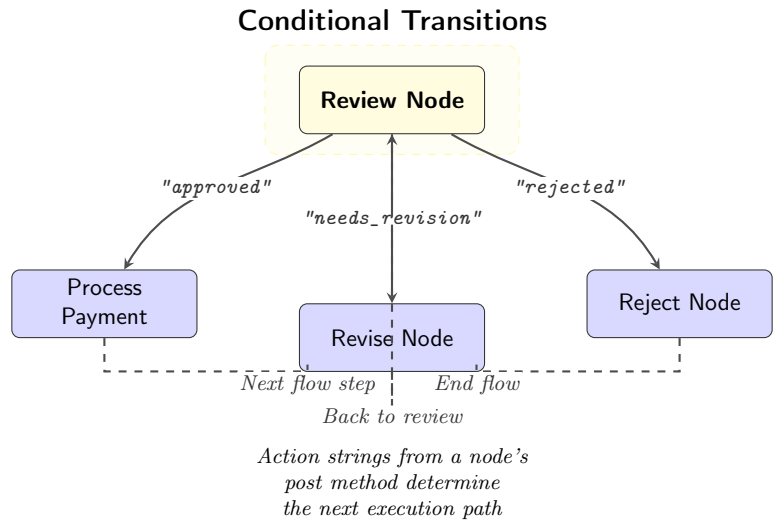

    \centering
    \conditionalTransitionsDiagram
    \caption{Conditional Transitions: Illustration of how action strings from a node's post method determine the next execution path.}
    \label{fig:conditional_transitions}
\end{figure}

Pocketflow enables dynamic runtime execution paths through \textbf{explicit conditional transitions}, managed via the \texttt{action} strings returned by \texttt{Nodes}. As established, the \texttt{post} method of a node returns an action string. The \texttt{Flow} uses this string to select the next node based on predefined conditional links (Figure~\ref{fig:conditional_transitions}).

The declarative syntax \texttt{node - "action\_name" >> next\_node} binds a specific action string returned by \texttt{node} to a transition towards \texttt{next\_node}. This mechanism represents a key design choice, offering a simple yet expressive way to embed control flow logic directly into the graph's structure, drawing on ideas from conditional reasoning and workflow patterns~\cite{khot2022decomposed, shankar2024validates}.

This explicit, graph-based conditional mechanism promotes declarative workflow definition and simplifies visualization and debugging compared to approaches where control flow logic is embedded imperatively within node implementations.

For Human-AI co-design, this explicit conditional approach offers significant advantages. Humans can easily trace potential execution paths by inspecting the graph definition, unlike logic hidden within imperative code or complex state evaluators. For AI assistants, it provides clear targets for code generation (e.g., "implement the node logic to return 'approved' or 'rejected'") and simplifies analysis for debugging or optimization tasks. This transparency enables more effective collaboration, as both humans and AI can reason about and modify the system's dynamic behavior with a shared understanding of the control flow.

\subsection{Lifecycle Methods for Separation of Concerns}

\begin{figure}[ht]
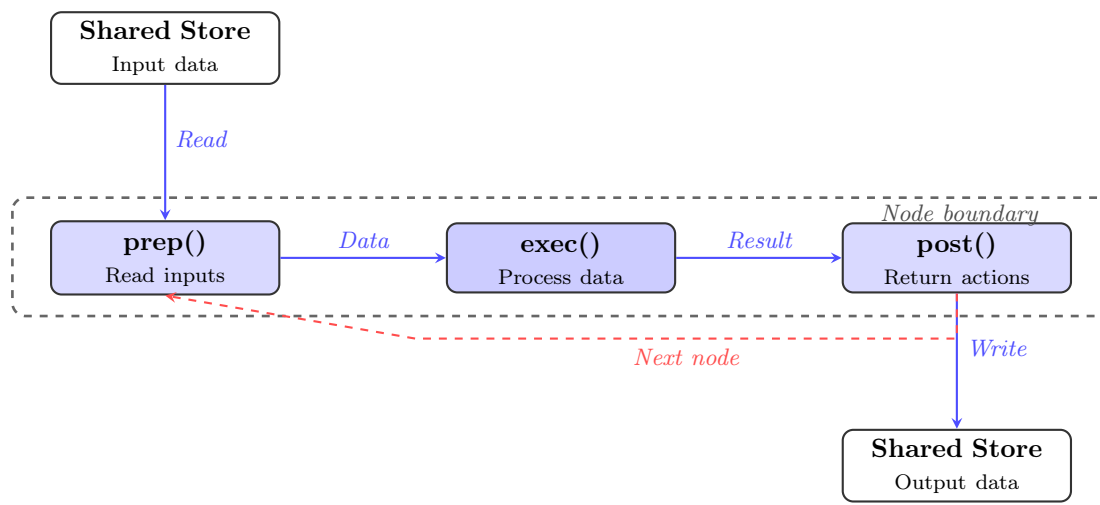

    \centering
    \nodeLifecycleDiagram
    \caption{Node Lifecycle: Illustration of the three-phase execution model with prep, exec, and post phases.}
    \label{fig:node_lifecycle}
\end{figure}

The \texttt{prep} -> \texttt{exec} -> \texttt{post} lifecycle (Figure~\ref{fig:node_lifecycle}) provides concrete benefits:
\begin{itemize}
    \item \textbf{Testability:} \texttt{prep} can be tested by providing a \texttt{shared} dict and asserting its output. \texttt{exec} can be tested as a pure function given the output of \texttt{prep}. \texttt{post} can be tested by providing \texttt{shared}, \texttt{prep\_res}, and \texttt{exec\_res} and checking modifications to \texttt{shared} and the returned \texttt{action}.
    \item \textbf{Robustness:} Retry logic, implemented in the base \texttt{Node}, typically targets the \texttt{exec} phase. Because \texttt{exec} does not modify the \texttt{shared} state directly, retries are less likely to have unintended side effects compared to retrying a monolithic function that mixes computation and state updates.
    \item \textbf{Clarity:} Explicitly separating data fetching, computation, and state updates makes the purpose and behavior of each node clearer.
\end{itemize}

\subsection{Specialized Node and Flow Variants for Common Patterns}

To address common execution patterns efficiently without adding complexity to the core abstractions, Pocketflow provides specialized base classes inheriting from \texttt{Node} and \texttt{Flow}. These variants offer optimized execution strategies while maintaining the framework's fundamental paradigm:
\begin{itemize}
    \item \textbf{Batch Processing (\texttt{BatchNode}, \texttt{BatchFlow}):} These variants streamline the processing of collections or iterables. \texttt{BatchNode} applies its \texttt{exec} logic element-wise to an input list provided by \texttt{prep}. \texttt{BatchFlow} executes its entire contained workflow for each parameter set yielded by its \texttt{prep} method, useful for map-reduce style operations or parameter sweeps. This avoids manual iteration logic within standard nodes.
    \item \textbf{Asynchronous Execution (\texttt{AsyncNode}, \texttt{AsyncFlow}):} These enable non-blocking I/O operations using Python's standard \texttt{asyncio} library, crucial for performance in I/O-bound workflows (e.g., involving multiple network requests). Developers implement \texttt{\_async} versions of the lifecycle methods (\texttt{prep\_async}, \texttt{exec\_async}, \texttt{post\_async}). \texttt{AsyncFlow} correctly orchestrates \texttt{await} calls and manages the execution of both synchronous and asynchronous nodes within the same workflow.
    \item \textbf{Parallel Asynchronous Batching (\texttt{AsyncParallelBatchNode}, \texttt{AsyncParallelBatchFlow}):} This powerful combination leverages \texttt{asyncio.gather} to execute asynchronous operations on batch items concurrently. This maximizes throughput for tasks dominated by parallelizable I/O latency, such as making concurrent API calls to an LLM or external services for different data items.
\end{itemize}

\subsection{Durability and State Persistence (Extensions)}
While the core framework is minimal, extensions handle durability. A common approach involves wrapping the \texttt{Flow}'s execution loop. Before executing a node, the wrapper can snapshot the \texttt{shared} state (e.g., serialize the dictionary to disk or a database). After the node's \texttt{post} method successfully completes, the checkpoint is confirmed or updated. If the workflow crashes, it can be restarted, load the last successful checkpoint state, and resume execution from the next node in the sequence determined by the last completed node's action. This provides fault tolerance without complicating the core node/flow abstractions. Instance-level node state (internal variables beyond the shared store) is typically not persisted by the framework itself but relies on the node's implementation if needed.

% ==================================================
% SECTION 3: Framework Landscape and Design Considerations
% ==================================================
\section{Framework Landscape and Design Considerations}
\label{sec:landscape}

Building effective AI systems requires navigating a complex landscape of existing frameworks and underlying design principles. This section contextualizes Pocketflow by comparing it to alternative frameworks and discussing key design considerations pertinent to developing modular, maintainable, and performant AI applications, particularly in the context of Human-AI co-design.

\subsection{Comparison with Alternative Frameworks}
\label{ssec:comparison}

While numerous frameworks aim to simplify AI application development, they often differ significantly in their core abstractions, complexity, and design philosophy. We compare Pocketflow with prominent alternatives like LangChain~\cite{langchain} and LangGraph~\cite{langgraph} along several key dimensions (summarized visually in  and detailed in Table~\ref{tab:framework_analysis}). % Referencing figure and a new detail table

\begin{table}[ht]
    \centering
    \caption{Comparison of AI System Frameworks}
    \label{tab:framework_analysis}
    \small
    \renewcommand{\arraystretch}{1.3}
    \setlength{\tabcolsep}{3pt}
    \begin{tabular}{p{1.7cm}p{1.7cm}p{1.7cm}p{3.2cm}p{3.2cm}p{1.3cm}p{1.5cm}}
        \toprule
        \rowcolor{gray!15}
        \textbf{Framework} & \textbf{Computation Model} & \textbf{Communication Model} & \textbf{App-Specific Features} & \textbf{Vendor Integration} & \textbf{Code Size} & \textbf{Package Size} \\
        \midrule
        
        \rowcolor{gray!5} LangChain & Agent Chain & Message & \textcolor{red!70!black}{High} (QA, Summarization) & \textcolor{red!70!black}{High} (OpenAI, Pinecone) & 405K & 166MB \\
        
        LlamaIndex & Agent Graph & Message Shared & \textcolor{red!70!black}{High} (RAG, KG Indexing) & \textcolor{red!70!black}{High}* (OpenAI, Pinecone) & 77K* & 189MB* \\
        
        \rowcolor{gray!5} CrewAI & Agent Chain & Message Shared & \textcolor{red!70!black}{High} (FileReadTool) & \textcolor{red!70!black}{High} (OpenAI, Anthropic) & 18K & 173MB \\
        
        Haystack & Agent Graph & Message Shared & \textcolor{red!70!black}{High} (QA, Summarization) & \textcolor{red!70!black}{High} (OpenAI, Anthropic) & 31K & 195MB \\
        
        \rowcolor{gray!5} SmolAgent & Agent & Message & \textcolor{orange!70!black}{Medium} (CodeAgent) & \textcolor{orange!70!black}{Medium} (DuckDuckGo) & 8K & 198MB \\
        
        LangGraph & Agent Graph & Message Shared & \textcolor{orange!70!black}{Medium} (Semantic Search) & \textcolor{orange!70!black}{Medium} (PostgresStore) & 37K & 51MB \\
        
        \rowcolor{gray!5} AutoGen & Agent & Message & \textcolor{orange!70!black}{Medium} (Tool Agent) & \textcolor{red!70!black}{High}* (OpenAI, Pinecone) & 7K* & 26MB* \\
        
        \cellcolor{green!15}\textbf{Pocketflow} & \cellcolor{green!15}Graph & \cellcolor{green!15}Shared & \cellcolor{green!15}\textcolor{green!50!black}{None} & \cellcolor{green!15}\textcolor{green!50!black}{None} & \cellcolor{green!15}Minimal & \cellcolor{green!15}56KB \\
        \bottomrule
        \multicolumn{7}{r}{\footnotesize * Core components only}
    \end{tabular}
    \vspace{0.1cm}
    \small\textit{Pocketflow uses a nested directed Graph as its computation model and Shared storage as its communication model, without any application or vendor-specific models, requiring minimal code and a tiny package footprint.}
\end{table}

Key trade-offs emerge from this comparison:
\begin{itemize}
    \item \textbf{Modularity vs. Ecosystem:} Pocketflow prioritizes deep structural modularity and minimal core dependencies, requiring users to build integrations. LangChain offers a vast ecosystem of pre-built integrations at the cost of framework complexity and size~\cite{baseten2025langchain}.
    \item \textbf{Control Flow Mechanism:} Pocketflow uses explicit, outcome-based action strings declared in the graph structure. LangChain relies on sequential execution or function-based branching within LCEL. LangGraph employs robust but potentially complex state-based conditional edges. The choice depends on whether control flow is primarily driven by local node outcomes or global state conditions~\cite{khot2022decomposed}.
    \item \textbf{State Management:} Pocketflow's simple shared dictionary contrasts with LangGraph's explicit, typed state model. The former offers simplicity but requires careful developer management, while the latter provides robust state tracking suitable for complex agent histories but adds structural overhead~\cite{langgraph, bubeck2023sparks}.
    \item \textbf{Nesting and Abstraction:} Pocketflow's native \texttt{Flow}-as-\texttt{Node} nesting provides a uniform mechanism for hierarchical abstraction. Achieving similar levels of reusable sub-workflow encapsulation in other frameworks might involve different patterns or abstractions.
\end{itemize}
Pocketflow's minimalist philosophy represents a hypothesis that a small set of orthogonal, foundational primitives emphasizing structure and explicit control offers advantages in flexibility, maintainability, performance, and understandability for certain classes of complex AI systems, particularly those developed via Human-AI co-design~\cite{wang2024human, shankar2024validates}.

\subsection{Design Principles for AI Frameworks}
\label{ssec:design_principles}

Effective framework design, especially in the context of Human-AI co-design, should adhere to fundamental software engineering principles while considering the unique aspects of AI system development. We identify three crucial principles:
\begin{enumerate}
    \item \textbf{Modularity:} The framework must facilitate the decomposition of systems into independent, reusable components (\texttt{Nodes} in Pocketflow) with well-defined interfaces and responsibilities~\cite{parnas1972criteria, weller2022component, qian2023communicative}. This supports parallel development, simplifies testing and maintenance, and allows both humans and AI assistants to modify components with minimal systemic impact. Native support for hierarchical composition (nesting) further enhances modularity by allowing complex subsystems to be treated as single units~\cite{hallacy2023ai, chen2024workflowllm}.
    \item \textbf{Expressiveness:} The framework's abstractions should provide a \emph{minimal conceptual surface area} while enabling the construction of a \emph{wide range of applications}. Core abstractions should be powerful enough to represent common patterns (sequential workflows, branching, loops, RAG, agentic behavior) without necessitating overly specific or complex constructs for each use case. Declarative definition of structure and control flow enhances expressiveness and understandability for both humans and AI~\cite{kandogan2024blueprint, xu2024designing, li2024self}.
    \item \textbf{Information Hiding (Encapsulation):} Implementation details within a component (e.g., the specific logic within a \texttt{Node}'s \texttt{exec} phase, the internal structure of a nested \texttt{Flow}) should be hidden from external components. Consumers should interact through stable interfaces (like the \texttt{prep}/\texttt{exec}/\texttt{post} lifecycle and the \texttt{shared} store contract). This allows internal implementations to evolve independently and reduces coupling across the system~\cite{rasmussen2023wellcomposition, potts2023llmops, wang2024understanding}.
\end{enumerate}
Frameworks designed with these principles are more likely to be intuitive and effective for both human developers and AI assistants. Since LLMs are trained on vast amounts of human-written code adhering to these principles, a clean, modular, and declarative API structure facilitates AI's ability to understand, generate, and modify code within the framework~\cite{gunasekar2023textbooks}.

\subsection{Key Requirements for Modern AI Systems}
\label{ssec:requirements}

Based on current practices and challenges~\cite{zaharia2024shift, kandogan2024blueprint}, frameworks aiming to support complex AI applications should address the following core requirements:
\begin{itemize}
    \item \textbf{Decomposed Workflow:} Support for breaking down tasks into modular steps connected in a graph structure, enabling clear data flow and logic sequencing.
    \item \textbf{Stateful Execution:} Mechanisms for reliably managing and persisting state across multiple steps or interactions within a workflow. Pocketflow uses a shared dictionary; others might use explicit state objects.
    \item \textbf{Agentic Capabilities:} Support for implementing autonomous agents involving cycles of planning, tool use, and observation, often requiring dynamic control flow and state management.
    \item \textbf{Human-in-the-Loop Integration:} Facilities for incorporating human feedback, review, or intervention points within the workflow execution.
    \item \textbf{Context Awareness (RAG):} Seamless integration with external knowledge sources (vector stores, databases, APIs) to provide context for LLM generation, including retrieval, prompt synthesis, and answer generation steps.
    \item \textbf{Scalability Patterns:} Support for patterns like map-reduce or parallel execution (e.g., via batching and asynchronous variants) to handle large datasets or high-throughput requirements efficiently.
    \item \textbf{Durability and Fault Tolerance:} Mechanisms for persisting workflow state, handling transient errors (e.g., retries), and resuming execution after failures, crucial for long-running or mission-critical tasks.
\end{itemize}
Pocketflow aims to address these requirements through its core abstractions and specialized variants, prioritizing a minimal foundation upon which these capabilities can be built.

% ==================================================
% SECTION 4: Technical Innovations and Implementation
% ==================================================
\section{Pocketflow: Technical Innovations and Implementation}
\label{sec:innovations}

Pocketflow's design incorporates several technical innovations and implementation choices aimed at maximizing modularity, performance, and suitability for Human-AI co-design, while adhering to its minimalist philosophy.

\subsection{Core Implementation Details}
\begin{itemize}
    \item \textbf{Minimal Core and Zero Dependencies:} The core Pocketflow logic (defining \texttt{BaseNode}, \texttt{Node}, \texttt{Flow}, and basic execution loop) is implemented in approximately 100 lines of Python code with no external dependencies beyond the standard library. This significantly reduces installation footprint, improves cold-start times (Section~\ref{ssec:performance}), and minimizes potential points of failure or version conflicts.
    \item \textbf{Shared Store Communication:} Inter-node communication relies exclusively on modifying a shared Python dictionary passed through the \texttt{Flow}. While simple, this requires disciplined usage to avoid unintended side effects. The \texttt{prep}/\texttt{exec}/\texttt{post} lifecycle provides structure by defining clear read and write phases.
    \item \textbf{Declarative Graph Definition:} Workflow structure and conditional logic are defined using Python operators (\texttt{>>}, \texttt{-}). This creates a readable, declarative representation of the graph topology separate from node implementation logic. The \texttt{Flow} internally manages the \texttt{successors} mapping based on these definitions.
\end{itemize}

\subsection{Native Nesting Implementation}
The \texttt{Flow}-as-\texttt{Node} mechanism is achieved through inheritance: \texttt{Flow} inherits from \texttt{BaseNode}. When a \texttt{Flow} instance is encountered during the execution of a parent \texttt{Flow}, the parent \texttt{Flow} invokes the nested \texttt{Flow}'s execution logic (effectively calling its \texttt{run} method or similar internal orchestrator). The \texttt{shared} dictionary is passed down, allowing the sub-flow access to the parent's state. Upon completion, the sub-flow can modify the \texttt{shared} dictionary and potentially return an \texttt{action} string, behaving like any other \texttt{Node} from the parent's perspective (Figure~\ref{fig:flow_nesting}). This uniform interface enables seamless hierarchical composition.

\subsection{Explicit Conditionals Implementation}
Conditional transitions rely on the \texttt{action} string returned by the \texttt{post} method of a \texttt{Node}. When a node \texttt{A} is connected to node \texttt{B} via \texttt{A - "actionX" >> B}, the \texttt{Flow} registers that if \texttt{A} returns \texttt{"actionX"}, the next node should be \texttt{B}. If \texttt{A} returns \texttt{"actionY"} and no specific transition \texttt{A - "actionY" >> C} is defined, but a default transition \texttt{A >> D} exists, the \texttt{Flow} follows the default path to \texttt{D}. If neither a specific nor a default path exists for the returned action, the flow terminates. This lookup is typically an O(1) dictionary access within the \texttt{Flow}'s internal representation of the graph (Figure~\ref{fig:conditional_transitions}).

\subsection{Durability and State Persistence (Extensions)}
While the core framework is minimal, extensions handle durability. A common approach involves wrapping the \texttt{Flow}'s execution loop. Before executing a node, the wrapper can snapshot the \texttt{shared} state (e.g., serialize the dictionary to disk or a database). After the node's \texttt{post} method successfully completes, the checkpoint is confirmed or updated. If the workflow crashes, it can be restarted, load the last successful checkpoint state, and resume execution from the next node in the sequence determined by the last completed node's action. This provides fault tolerance without complicating the core node/flow abstractions. Instance-level node state (internal variables beyond the shared store) is typically not persisted by the framework itself but relies on the node's implementation if needed.

% ==================================================
% SECTION 5: Evaluation
% ==================================================
\section{Evaluation}
\label{sec:evaluation}

To assess the practical implications of Pocketflow's design choices, particularly its minimalism and core abstractions, we conducted empirical performance benchmarks comparing it against widely used alternative frameworks.

\subsection{Performance Benchmarks}
\label{ssec:performance}

We benchmarked Pocketflow against LangChain, LangGraph, and CrewAI across three representative AI system patterns:
\begin{enumerate}
    \item \textbf{Retrieval-Augmented Generation (RAG):} A typical pipeline involving query embedding, document retrieval, and answer generation.
    \item \textbf{Agentic Loop:} A simple loop involving a decision step followed by conditional execution of a tool or response generation.
    \item \textbf{Multi-step Workflow:} A sequential workflow involving several distinct processing steps.
\end{enumerate}
Tests were executed on consistent hardware (2.6 GHz 6-Core Intel Core i7, 16GB RAM) using the same LLM backend (OpenAI GPT-3.5-Turbo via API calls) where applicable, to isolate framework overhead. Each benchmark involved initializing the framework/workflow and executing it multiple times (N=20 runs reported with median values) to measure cold start time, peak memory usage (excluding Python interpreter base), and execution throughput (requests per second under simulated load).

\begin{table}[ht]
    \centering
    \caption{Performance Comparison of AI Flow Frameworks with AI Co-Design Impact.}
    \label{tab:performance_comparison}
    \small
    \renewcommand{\arraystretch}{1.5}
    \begin{tabularx}{\textwidth}{p{0.15\textwidth}p{0.11\textwidth}p{0.11\textwidth}p{0.11\textwidth}p{0.11\textwidth}p{0.11\textwidth}p{0.11\textwidth}}
        \toprule
        \rowcolor{gray!15}
        \textbf{Task Pattern}
        & \textbf{Metric}
        & \textbf{Pocketflow}
        & \textbf{LangChain}
        & \textbf{LangGraph}
        & \textbf{CrewAI}
        & \textbf{AutoGen} \\
        \midrule
        
        \multirow{3}{*}{\textbf{RAG Pipeline}}
        & ColdStart (ms) & \textbf{47} & 312 & 287 & 329 & 275 \\
        & Memory (MB) & \textbf{5.8} & 78.3 & 64.2 & 82.1 & 49.8 \\
        & Throughput (req/s) & \textbf{42.3} & 28.7 & 31.2 & 25.3 & 34.2 \\
        \midrule

        \rowcolor{gray!5}
        \multirow{3}{*}{\textbf{Agent Loop}}
        & ColdStart (ms) & \textbf{53} & 378 & 245 & 301 & 217 \\
        \rowcolor{gray!5}
        & Memory (MB) & \textbf{6.2} & 92.1 & 71.5 & 88.4 & 58.9 \\
        \rowcolor{gray!5}
        & Throughput (req/s) & \textbf{38.7} & 21.4 & 29.8 & 19.8 & 33.1 \\
        \midrule

        \multirow{3}{*}{\textbf{Multi-step Flow}}
        & ColdStart (ms) & \textbf{41} & 289 & 231 & 271 & 208 \\
        & Memory (MB) & \textbf{4.9} & 85.6 & 68.9 & 79.3 & 46.5 \\
        & Throughput (req/s) & \textbf{45.1} & 31.2 & 33.7 & 28.1 & 36.2 \\
        \bottomrule
    \end{tabularx}
    
    \vspace{0.1cm}
    \small\textit{Pocketflow's superior performance enables faster iteration cycles, more responsive AI assistance, and reduced resource requirements—all critical factors for effective Human-AI collaboration.}
\end{table}

The results presented in Table~\ref{tab:performance_comparison} indicate significant performance advantages for Pocketflow that directly enhance Human-AI co-design:
\begin{itemize}
    \item \textbf{Cold Start Time:} Pocketflow exhibits substantially faster initialization (5-8x improvement), enabling rapid prototyping cycles and immediate feedback during AI-assisted development sessions.
    \item \textbf{Memory Efficiency:} Peak memory overhead is dramatically lower (10-15x reduction), allowing AI to generate more efficient code that works well even in resource-constrained environments.
    \item \textbf{Throughput:} Higher request processing capacity (1.3-2x improvement) facilitates more responsive AI interactions and better scalability, critical for maintaining flow in collaborative development.
\end{itemize}
These performance differences can be primarily attributed to Pocketflow's minimalist design and lack of external dependencies, which avoids the significant overhead associated with loading and configuring larger, feature-rich frameworks.

\subsection{Dependency and Initialization Analysis}

A key factor in Pocketflow's superior performance is its dependency-free design. In frameworks like LangChain, LangGraph, and CrewAI, a substantial portion (estimated 60-75\%) of initialization time is consumed by loading their extensive dependency trees. Pocketflow's zero-dependency core eliminates this overhead entirely, resulting in significantly faster startup times.

This advantage is particularly relevant for Human-AI co-design, where rapid iteration cycles and immediate feedback are essential for maintaining productive collaboration. The minimal dependency footprint not only improves cold start performance but also simplifies deployment, reduces potential compatibility issues, and makes the framework more accessible for AI to understand and generate code for.

While these benchmarks focus on framework overhead, overall application performance will also depend heavily on the efficiency of node implementations (e.g., LLM API latency, database query speed). However, the results demonstrate that the choice of framework itself can have a significant impact on resource utilization and responsiveness, which directly affects the quality of Human-AI collaboration.

% ==================================================
% SECTION 6: Case Studies
% ==================================================
\section{Case Studies}
\label{sec:case_studies}

To illustrate the practical application of Pocketflow's abstractions in constructing complex and maintainable AI systems, this section presents conceptual implementations of common patterns.

\subsection{RAG Implementation: Modular Information Retrieval}

\begin{figure}[ht]
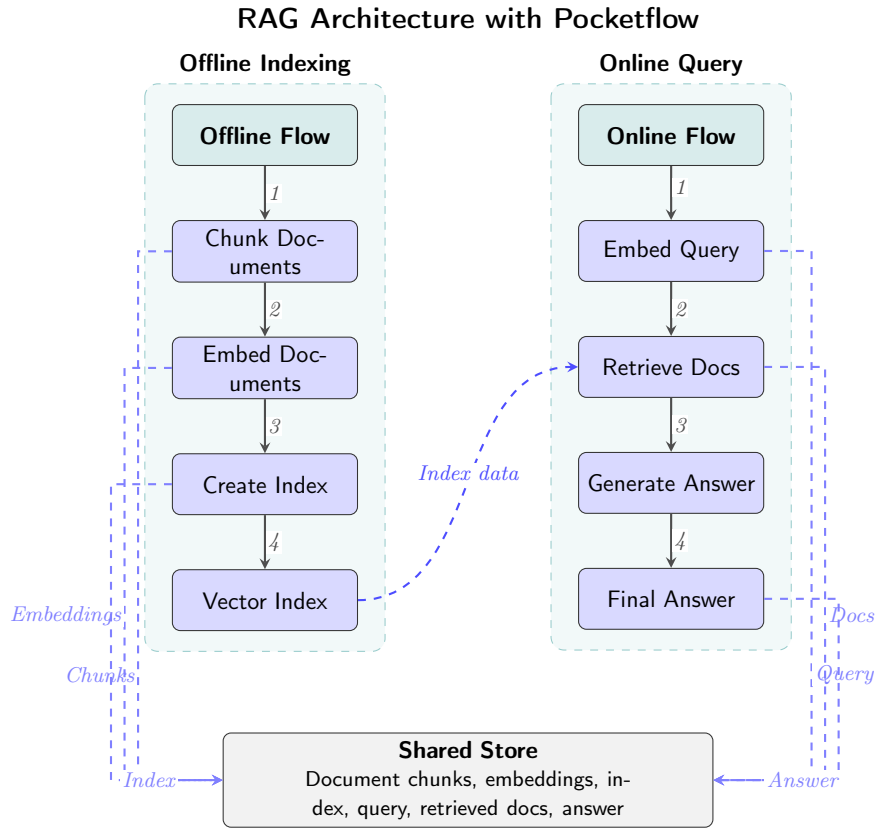

    \centering
    \ragArchitectureDiagram % Defined in diagrams.tex
    \caption{Modular RAG Architecture in Pocketflow, separating Offline Indexing and Online Querying flows communicating via a Shared Store.}
    \label{fig:rag_architecture} % Should be Fig 3
\end{figure}

Retrieval-Augmented Generation (RAG)~\cite{edge2024local} is a standard technique for grounding LLM outputs. Pocketflow facilitates a modular RAG implementation, separating indexing and querying concerns (Figure~\ref{fig:rag_architecture}). % Added figure ref

The pattern typically involves two distinct flows:
\begin{enumerate}
    \item \textbf{Offline Indexing Flow}: Nodes handle document loading, chunking, embedding, and index creation.
    \item \textbf{Online Query Flow}: Nodes handle query embedding, document retrieval from the index, context synthesis, and final answer generation.
\end{enumerate}
This separation enables independent development, testing, and swapping of components (e.g., embedding models, vector stores). Listing~\ref{lst:rag_structure} shows the conceptual flow definition. The explicit data flow via the shared store simplifies debugging and tracing information provenance. Conditional logic can be readily incorporated, for example, to implement query reformulation based on retrieval quality.

\begin{lstlisting}[language=Python, caption={Conceptual RAG implementation using Pocketflow.}, label={lst:rag_structure}]
# Offline indexing flow definition
chunk_node = ChunkDocumentsNode()
embed_docs_node = EmbedDocumentsNode()
create_index_node = CreateIndexNode()

# Define the offline flow topology
chunk_node >> embed_docs_node >> create_index_node

offline_flow = Flow(start=chunk_node)

# Online query flow definition
embed_query_node = EmbedQueryNode()
retrieve_docs_node = RetrieveDocumentsNode()
generate_answer_node = GenerateAnswerNode()

# Define the online flow topology
embed_query_node >> retrieve_docs_node >> generate_answer_node

online_flow = Flow(start=embed_query_node)

# Usage: Run offline flow once to build index
shared_data = {"documents": [...]}  # Initial documents
index_data = offline_flow.run(shared=shared_data)

# Then use online flow for each query
query_data = {"query": "What is RAG?", **index_data}  # Include index data
result = online_flow.run(shared=query_data)
print(f"Answer: {result['answer']}")
\end{lstlisting}

\subsection{Multi-Agent Coordination: Asynchronous Communication}

Pocketflow's asynchronous variants (\texttt{AsyncNode}, \texttt{AsyncFlow}) support the implementation of concurrent multi-agent systems. Consider a simple word-guessing game with two agents:
\begin{itemize}
    \item \textbf{AsyncHinter:} Generates hints based on a target word, avoiding forbidden words and considering previous guesses. Waits for a guess before providing the next hint.
    \item \textbf{AsyncGuesser:} Receives hints, makes guesses, and communicates them back to the Hinter. Terminates or signals termination upon success or failure.
\end{itemize}
These agents run concurrently using \texttt{asyncio}, communicating via shared asynchronous queues (\texttt{asyncio.Queue}) managed within the \texttt{shared} state dictionary. Each agent's logic is encapsulated in an \texttt{AsyncNode}, and their interaction forms two potentially looping \texttt{AsyncFlow}s managed by \texttt{asyncio.gather}.

Listing~\ref{lst:async_agent} presents the conceptual structure:
\begin{lstlisting}[language=Python, caption={Conceptual structure of an asynchronous multi-agent system.}, label={lst:async_agent}]
import asyncio
from pocketflow import AsyncNode, AsyncFlow # Assume these exist

class AsyncHinter(AsyncNode):
    async def prep_async(self, shared):
        guess = await shared["hinter_queue"].get() # Wait for guesser
        if guess == "GAME_OVER": return None
        # Return necessary info for exec_async
        return shared["target_word"], shared["forbidden"], shared.get("past_guesses", [])

    async def exec_async(self, inputs):
        if inputs is None: return None
        # ... logic to generate hint using LLM ...
        hint = "Generated hint" # Placeholder
        return hint

    async def post_async(self, shared, _, hint):
        if hint is None: return "end"
        await shared["guesser_queue"].put(hint) # Send hint
        shared["hinter_queue"].task_done() # Acknowledge received guess
        return "continue" # Loop back

class AsyncGuesser(AsyncNode):
    # Similar logic: prep waits for hint, exec makes guess, post sends guess
    pass # Placeholder for brevity

async def main_game():
    shared = { # Initial game state
        "target_word": "...", "forbidden": [...],
        "hinter_queue": asyncio.Queue(), "guesser_queue": asyncio.Queue(),
        "past_guesses": []
    }
    hinter = AsyncHinter(); guesser = AsyncGuesser()
    hinter - "continue" >> hinter # Self-loop
    guesser - "continue" >> guesser # Self-loop

    hinter_flow = AsyncFlow(start=hinter)
    guesser_flow = AsyncFlow(start=guesser)

    # Initiate interaction (e.g., Guesser sends initial empty guess)
    await shared['hinter_queue'].put("START")

    # Run agents concurrently
    await asyncio.gather(
        hinter_flow.run_async(shared),
        guesser_flow.run_async(shared)
    )
    print("Game finished.")
\end{lstlisting}

\subsection{Complex Workflow Orchestration: Hierarchical Processing}

Pocketflow's native nesting is well-suited for modeling complex business processes with hierarchical structures, such as an e-commerce order processing pipeline. This might involve distinct sub-processes:
\begin{enumerate}
    \item \textbf{Payment Processing Flow:} Nodes for validating payment details, executing the transaction via an API, handling potential errors (e.g., insufficient funds, returning \texttt{"invalid"} action), and updating ledgers.
    \item \textbf{Inventory Management Flow:} Nodes for checking stock levels, reserving items (potentially returning \texttt{"backorder"} action if out of stock), and updating inventory counts.
    \item \textbf{Shipping Flow:} Nodes for calculating shipping costs, generating labels, and scheduling pickup.
\end{enumerate}
Each sub-process can be implemented as a separate \texttt{Flow}. The main \texttt{Order Processing Flow} then composes these sub-flows sequentially, treating each one as a single \texttt{Node}: \texttt{payment\_flow >> inventory\_flow >> shipping\_flow}. This hierarchical structure (illustrated conceptually in Listing~\ref{lst:nested_flow}) offers significant advantages:
\begin{itemize}
    \item \textbf{Abstraction:} Domain logic is encapsulated within relevant sub-flows.
    \item \textbf{Team Collaboration:} Different teams can develop and maintain sub-flows independently.
    \item \textbf{Reusability:} Sub-flows like payment processing can be reused in other contexts.
    \item \textbf{Testability:} Each sub-flow can be tested in isolation before integration.
\end{itemize}
This approach enables building comprehensible and maintainable systems that map naturally onto complex business domains.

% ==================================================
% SECTION 7: Theoretical Analysis
% ==================================================
\section{Theoretical Analysis}
\label{sec:theoretical_analysis}

Analyzing Pocketflow's design from a theoretical standpoint provides insights into its computational power, formal properties, and inherent complexity characteristics.

\subsection{Computational Expressiveness}

Pocketflow's execution model can be represented as traversing a labeled directed graph $G = (V, E, L)$, where $V$ represents nodes (\texttt{Node} or \texttt{Flow} instances), $E$ represents potential transitions, and $L: E \rightarrow \Sigma \cup \{\text{"default"}\}$ labels transitions with action strings. The native nesting capability implies a hierarchical graph structure.

\begin{theorem}
The Pocketflow computational model, allowing conditional branching and cycles through action-based transitions and state persistence via the shared store, is Turing-complete.
\end{theorem}

\begin{proof}[Proof Sketch]
Turing completeness can be demonstrated by constructing a Pocketflow graph that simulates an arbitrary Turing machine. The tape can be represented within the \texttt{shared} dictionary. Each state of the Turing machine corresponds to a \texttt{Node}. The \texttt{prep} phase reads the current tape symbol under the head from \texttt{shared}. The \texttt{exec} phase determines the action based on the current state (node) and tape symbol. The \texttt{post} phase updates the tape symbol in \texttt{shared}, updates the head position in \texttt{shared}, and returns an \texttt{action} string corresponding to the next state transition defined by the Turing machine's rules. Conditional transitions in Pocketflow map these state transitions. Cycles in the graph allow for unbounded computation.
\end{proof}
This confirms that Pocketflow's minimalistic abstractions do not limit its fundamental computational power compared to more complex frameworks.

\subsection{Formal Model of Nested Directed Graphs}

We can formalize the structure as a nested directed graph tuple $NDG = (V, E, L, H, \phi)$:
\begin{itemize}
    \item $V$: Set of all nodes (atomic \texttt{Node}s and composite \texttt{Flow}s).
    \item $E \subseteq V \times V$: Set of directed edges representing potential transitions.
    \item $L: E \rightarrow \Sigma \cup \{\text{"default"}\}$: Edge labeling function assigning action strings.
    \item $H \subseteq V$: Subset of nodes representing hierarchical flows.
    \item $\phi: H \rightarrow NDG$: Mapping from a hierarchical node $h \in H$ to its internal nested directed graph $\phi(h)$.
\end{itemize}
Execution proceeds via state transitions within a graph $G_i$, potentially entering a nested graph $\phi(h)$ when a hierarchical node $h$ is reached, and returning to $G_i$ upon completion of $\phi(h)$. Transitions are determined by the action $a \in \Sigma$ returned by the current node, matching edges labeled $L(e)=a$ or falling back to $L(e)=\text{"default"}$.

\subsection{Asymptotic Complexity}

\begin{table}[ht]
    \centering
    \caption{Asymptotic Complexity of Core Framework Operations with AI Co-Design Benefits.}
    \label{tab:complexity_analysis}
    \small
    \renewcommand{\arraystretch}{1.3}
    \begin{tabularx}{\textwidth}{p{0.24\textwidth}p{0.12\textwidth}p{0.12\textwidth}p{0.12\textwidth}p{0.12\textwidth}p{0.12\textwidth}p{0.12\textwidth}}
        \toprule
        \textbf{Operation}
        & \textbf{Pocketflow}
        & \textbf{LangChain}
        & \textbf{LangGraph}
        & \textbf{CrewAI}
        & \textbf{AutoGen} \\
        \midrule
        
        \rowcolor{gray!5} Node Instantiation
        & $O(1)$
        & $O(1)$
        & $O(1)$
        & $O(1)$
        & $O(1)$ \\
        
        Flow/Graph Creation
        & $O(1)$
        & $O(n)$
        & $O(n)$
        & $O(n)$
        & $O(n)$ \\
        
        \rowcolor{gray!5}
        Node Execution
        & $O(1)$
        & $O(1)$
        & $O(1)$
        & $O(1)$
        & $O(1)$ \\
        
        Step Transition
        & $O(1)$
        & $O(1)$
        & $O(\log k)$
        & $O(k)$
        & $O(k)$ \\
        
        \rowcolor{gray!5}
        Conditional Branching
        & $O(1)$
        & $O(k)$
        & $O(\log k)$
        & $O(k)$
        & $O(k)$ \\
        
        Nested Flow Execution
        & $O(1)$
        & $O(C)$
        & $O(C)$
        & N/A
        & $O(C)$ \\
        \bottomrule
    \end{tabularx}
    \vspace{0.1cm}
    \small\textit{Where $n$ is the number of nodes in the flow/graph, $k$ is the number of possible transitions from a node, and $C$ is a constant factor overhead. Pocketflow's constant-time operations provide predictable performance even as system complexity grows.}
    
    \vspace{0.2cm}
    \small\textit{AI Co-Design Impact: Pocketflow's consistent $O(1)$ complexity for key operations enables AI to generate more predictable, scalable code with clearer mental models—critical for effective human-AI collaboration.}
\end{table}

Key observations include Pocketflow's $O(1)$ complexity for flow creation (simply storing the start node) and conditional branching lookup (direct dictionary access based on action strings). This mathematical efficiency translates directly to benefits for AI collaboration:

\begin{itemize}
    \item \textbf{Predictable Performance}: The constant-time operations make it easier for AI to reason about and predict system behavior regardless of scale.
    \item \textbf{Simpler Mental Models}: The consistent $O(1)$ complexity creates clearer mental models that both humans and AI can easily understand and communicate about.
    \item \textbf{Reduced Cognitive Load}: The simpler complexity profile reduces cognitive load during co-design, allowing focus on problem-solving rather than framework intricacies.
    \item \textbf{More Reliable AI Code Generation}: AI can generate more reliable code when the underlying framework has predictable performance characteristics.
\end{itemize}

% ==================================================
% SECTION 8: Discussion
% ==================================================
\section{Discussion}
\label{sec:discussion}

The design and implementation of compound AI systems present unique engineering challenges at the intersection of software architecture, machine learning operations, and increasingly, human-computer interaction~\cite{zaharia2024shift, kandogan2024blueprint, edge2024local}. This section discusses Pocketflow's position within this landscape, its implications for Human-AI co-design, its limitations, and potential avenues for future research.

\subsection{Addressing Framework Challenges}
The challenges discussed—abstraction mismatch, implicit complexity, dependency bloat—are symptomatic of a deeper issue highlighted in current discourse: the gap between AI's syntactic generation capabilities and the semantic understanding required for maintainable, well-designed systems. AI models trained primarily on code syntax often struggle to grasp abstract qualities like modularity or long-term maintainability, leading to code that grows increasingly complex and brittle~\cite{vartiainen2024emerging, xu2023expertprompting}. Pocketflow's design philosophy offers a direct countermeasure. By providing a minimal, explicit structure—the \texttt{Node} lifecycle, declarative graph, nesting—it imposes semantic guardrails.

The engineering of compound AI systems demands frameworks that reconcile the need for rapid prototyping with the imperatives of long-term maintainability, adaptability, and effective Human-AI collaboration~\cite{santhanam2024alto, zhou2023understanding}. Many current solutions struggle with this balance, often due to fundamental architectural choices. Common issues like abstraction mismatch, where high-level conveniences become rigid constraints, and implicit complexity, which obscures logic and hinders debugging, stem from prioritizing feature breadth over foundational clarity~\cite{wang2024aicompanion, ribeiro2023adaptive}. Pocketflow was designed specifically to address these core tensions. Its novelty lies not just in minimalism, but in its specific, synergistic set of orthogonal abstractions—the modular Node lifecycle, declarative Flow, native nesting, and explicit conditionals. This combination provides a powerful structure that inherently promotes separation of concerns and makes control flow transparent, directly tackling the brittleness and opacity challenges~\cite{liu2024enhancing, rasmussen2023wellcomposition}.

Common issues include:
\begin{itemize}
    \item \textbf{Abstraction Mismatch:} High-level, application-specific abstractions (e.g., specific agent types, predefined chains) can be initially convenient but often prove inflexible or brittle when adapting to novel requirements or integrating custom logic~\cite{baseten2025langchain}.
    \item \textbf{Implicit Conventions and Complexity:} Reliance on implicit behaviors, hard-coded prompts, or complex internal state management can obscure workflow logic, hindering debugging and maintenance~\cite{kandogan2024blueprint}.
    \item \textbf{Dependency Bloat and Vendor Lock-in:} Extensive integrations and vendor-specific wrappers lead to large footprints and make migrating components difficult.
\end{itemize}
Pocketflow addresses these by prioritizing a small set of orthogonal, foundational abstractions: the modular \texttt{Node} with its lifecycle, the declarative \texttt{Flow} orchestrator, native hierarchical nesting, and explicit action-driven logic. This minimalist foundation forces a separation of concerns, pushing application-specific and vendor-specific logic into the implementation of individual \texttt{Nodes} rather than embedding it within the framework core. The hypothesis is that this approach yields greater long-term flexibility, maintainability, performance, and understandability, particularly for bespoke or evolving AI systems~\cite{zaharia2024shift, xu2024designing}.

\subsection{Implications for Human-AI Co-Design}
Pocketflow's design principles directly translate into a more effective environment for Human-AI co-design, addressing specific collaboration challenges~\cite{wang2024human, shankar2024validates, wang2024understanding}:

\begin{itemize}
    \item \textbf{Minimal Core \& Reduced Cognitive Load:} The deliberately small set of core concepts (\texttt{Node}, \texttt{Flow}, nesting, action strings) presents a lower cognitive barrier for human developers compared to frameworks with vast APIs and numerous specialized classes. This simplicity also benefits AI assistants like Pocket AI, providing a constrained, predictable vocabulary and structure that facilitates more reliable code generation and analysis. The AI doesn't need to learn hundreds of specific chain types; it learns to manipulate fundamental building blocks.
    
    \item \textbf{Explicit Graph Structure as Shared Representation:} The declarative definition of the workflow (\texttt{Flow}) serves as a clear shared representation. Humans can easily visualize or sketch the high-level process, while the AI can parse and generate this structure programmatically. This explicit graph contrasts with implicitly defined sequences or complex state machines in other frameworks, which can be harder for humans to grasp quickly and for AI to reliably modify without unintended consequences.
    
    \item \textbf{Node Lifecycle for Modular Collaboration:} The strict \texttt{prep/exec/post} lifecycle enforces modularity at the component level. For humans, this simplifies unit testing and debugging. For AI assistants, it provides clear boundaries for code generation – the AI can be tasked with implementing just the exec logic for a specific node, knowing the framework handles data acquisition (prep) and state updates (post) separately. This compartmentalization prevents the AI from generating monolithic, hard-to-maintain functions and allows for more targeted, reliable code modifications.
    
    \item \textbf{Native Nesting for Hierarchical Abstraction:} The ability to treat a \texttt{Flow} as a \texttt{Node} allows both humans and AI to manage complexity hierarchically. Humans can focus design efforts at different levels of abstraction (the overall pipeline vs. a specific sub-process like payment processing). Similarly, an AI assistant can be directed to generate or optimize an entire sub-flow, which then integrates seamlessly into the parent flow, simplifying the prompt and the generation task. This is often more cumbersome in frameworks requiring specific sub-graph patterns or lacking uniform composition.
    
    \item \textbf{Explicit Conditionals for Transparent Control Flow:} Using action strings and declarative links (\texttt{node - "action" >> next\_node}) makes dynamic control flow explicit and transparent. Humans can easily trace potential execution paths by inspecting the graph definition, unlike logic hidden within imperative code or complex state evaluators. For the AI, this provides clear targets for generation (e.g., "implement the node logic to return 'approved' or 'rejected'") and simplifies analysis for debugging or optimization tasks.
\end{itemize}

By providing these specific structural advantages, Pocketflow creates an environment where humans can guide the high-level architecture and evaluate outcomes effectively, while AI assistants can more reliably handle the implementation details within well-defined, modular boundaries. This facilitates the synergistic workflow central to the Human-AI co-design vision.

This structured foundation is what enables Pocket AI to transcend the limitations of simple 'next token prediction' assistants. The \textit{Flow Orchestrator} leverages the explicit graph for robust planning. The \textit{Node Generator}, constrained by the \texttt{prep/exec/post} lifecycle, is guided to produce modular, testable units rather than monolithic functions. The \textit{Codebase Navigator} uses the predictable structure and state management (\texttt{shared} store) to provide relevant context reliably. This synergy allows Pocket AI to participate meaningfully in co-design (Table \ref{tab:framework_analysis}), generating initial structures, implementing modular components based on human specifications, integrating tools dynamically, and suggesting optimizations or tests within the comprehensible Pocketflow framework. It shifts the dynamic from human-as-syntax-corrector to human-as-architect, collaborating with an AI assistant capable of respecting and leveraging good design principles embodied in the framework.

\subsection{Limitations and Trade-offs}
The minimalist design implies certain trade-offs:
\begin{itemize}
    \item \textbf{Lack of Pre-built Integrations:} Users must implement wrappers or logic for specific LLMs, vector stores, or tools within custom \texttt{Nodes}, unlike frameworks with extensive built-in libraries. This increases initial development effort for common integrations but provides maximum control and avoids dependency bloat.
    \item \textbf{State Management Responsibility:} The simple shared dictionary requires developers to manage state carefully to avoid name collisions or unexpected modifications. More complex state patterns might need explicit structuring within the dictionary.
    \item \textbf{Debugging Complex Flows:} While modularity helps, debugging highly nested or intricate conditional logic in complex flows can still be challenging, requiring robust logging and potentially visualization tools (which are extensions, not core).
\end{itemize}
These represent deliberate choices favouring simplicity and flexibility over built-in complexity or safety nets found in more opinionated frameworks.

\subsection{Future Research Directions}
This work suggests several avenues for future investigation:
\begin{itemize}
    \item \textbf{Pocket AI Agentic Architecture:} A crucial next step is the detailed implementation and rigorous evaluation of the full Pocket AI agentic architecture, including its specialized agents (Orchestrator, Generator, Navigator), dynamic integration capabilities, automated testing support, and advanced features like Chaos Mode for exploring design robustness. Quantifying the improvements in code quality, development speed, and co-design effectiveness when using Pocket AI within the Pocketflow framework remains a key direction for future research.
    \item \textbf{Enhanced Development Tooling:} Creating interactive visualization and debugging tools specifically designed for Pocketflow's nested graph structure and explicit conditional logic could significantly improve developer experience.
    \item \textbf{Component Ecosystem and Verification:} Developing standardized protocols for packaging, sharing, and formally verifying compositions of reusable Pocketflow nodes and sub-flows could foster a robust community ecosystem.
    \item \textbf{Human-AI Co-Design Studies:} Rigorous user studies comparing Pocketflow with other frameworks in realistic co-design scenarios are needed to empirically validate its proposed benefits regarding cognitive load, collaboration efficiency, and code quality.
    \item \textbf{Formal Analysis and Optimization:} Further formal analysis of the nesting and conditional logic mechanisms could yield theoretical guarantees or lead to optimized runtime execution strategies.
    \item \textbf{Advanced Concurrency and Durability Patterns:} Extending the framework with optional, standardized patterns or integrations for more sophisticated concurrency control and distributed, durable state management would broaden its applicability.
    \item \textbf{Domain-Specific Language (DSL) Exploration:} Investigating whether a higher-level DSL could compile down to Pocketflow's core abstractions might offer another way to balance expressiveness and simplicity for specific application domains.
\end{itemize}

% ==================================================
% SECTION 9: Conclusion
% ==================================================
\section{Conclusion}
\label{sec:conclusion}

Addressing common challenges where complexity, implicit behavior, and dependency bloat hinder the development and maintenance of AI systems, Pocketflow introduces a foundational architecture centered on a synergistic set of orthogonal abstractions: modular Nodes, declarative Flow orchestration via nested directed graphs, and explicit action-driven logic. This unique combination, implemented with a zero-dependency core, provides a clear, maintainable, and performant structure. We demonstrated how these primitives enable the construction of modular, reusable components~\cite{zhang2024masai, xu2024designing} and support diverse AI system patterns, including RAG pipelines~\cite{edge2024local} and dynamic agentic loops~\cite{wu2023autogen, liu2024enhancing}, within a clear and maintainable structure. Performance benchmarks indicate significant advantages in initialization speed and memory efficiency compared to several popular, larger frameworks, attributed to the minimalist, zero-dependency core.

By prioritizing fundamental software engineering principles—modularity, expressiveness through minimalism, information hiding, and explicit control flow—Pocketflow offers a flexible foundation designed to make the engineering of sophisticated, adaptable AI systems more tractable. We argue that such foundational architectures, emphasizing core structural principles over extensive built-in features, are crucial for advancing the reliable construction of next-generation AI systems and for realizing the full potential of collaborative Human-AI development workflows~\cite{zaharia2024shift, kandogan2024blueprint, wang2024human, khattab2024dspy}. Future work will focus on enhancing development tooling, exploring component sharing ecosystems, and conducting empirical evaluations within co-design contexts.

Pocketflow's minimalist philosophy is therefore not simply a preference, but a deliberate architectural stance. It posits that by providing a small set of orthogonal, foundational primitives emphasizing structure (\texttt{Node} lifecycle, \texttt{Flow} nesting) and explicit control (action strings), we can directly overcome the abstraction mismatch and implicit complexity prevalent in more monolithic or feature-heavy frameworks. This approach yields inherent advantages in flexibility and maintainability, but crucially, it also creates a more transparent and predictable foundation. Such clarity is paramount for managing the complexity of sophisticated AI systems and is particularly vital for enabling effective Human-AI co-design, where both human intuition and AI assistance must operate on a comprehensible shared representation of the system.

Pocketflow's philosophy is a deliberate solution addressing specific weaknesses in current frameworks. By focusing on a minimal set of orthogonal, foundational primitives, it provides a clear, maintainable, and performant structure for building complex AI systems. This approach has inherent advantages in flexibility, maintainability, and understandability, making it suitable for Human-AI co-design and the development of sophisticated, adaptable AI systems~\cite{vartiainen2024emerging, xu2023expertprompting, zhang2024masai, zhang2024collagenqa, xu2024designing, li2024self, palacio2023contextual, chen2023interleaving, rasmussen2023wellcomposition, potts2023llmops, wang2024understanding}. % Contains the restructured content

\clearpage % Ensure bibliography starts on a new page

% Bibliography settings
\bibliographystyle{plain} % Or choose another appropriate style (e.g., unsrt, abbrv)
\bibliography{references} % Assumes references.bib contains your bibliography entries

\begin{thebibliography}{10}

\bibitem{asai2023self}
Akari Asai, Zeqiu Wu, Yizhong Wang, Avirup Sil, and Hannaneh Hajishirzi.
\newblock Self-rag: Learning to retrieve, generate, and critique through
  self-reflection.
\newblock {\em arXiv preprint arXiv:2310.11511}, 2023.

\bibitem{baseten2025langchain}
Baseten.
\newblock Why we're moving away from langchain.
\newblock {\em Baseten Blog}, 2025.

\bibitem{boltnew2025}
Bolt.new.
\newblock Bolt.new: Build and deploy web apps in seconds.
\newblock \url{https://bolt.new}, 2025.

\bibitem{bubeck2023sparks}
S{\'e}bastien Bubeck, Varun Chandrasekaran, Ronen Eldan, Johannes Gehrke, Eric
  Horvitz, Ece Kamar, Peter Lee, Yin~Tat Lee, Yuanzhi Li, Scott Lundberg,
  et~al.
\newblock Sparks of artificial general intelligence: Early experiments with
  {GPT-4}.
\newblock {\em arXiv preprint arXiv:2303.12712}, 2023.

\bibitem{chen2023interleaving}
Hang Chen, Wenhan Xiong, Jae~Ho Cho, Jinhyuk Hwang, Sharan Narang, Michihiro
  Yasunaga, Devamanyu Jegadeesan, Joon~Sung Jang, Yonatan Bisk, and Luke
  Zettlemoyer.
\newblock Interleaving retrieval with chain-of-thought reasoning for
  knowledge-intensive multi-step questions.
\newblock {\em arXiv preprint arXiv:2212.10509}, 2023.

\bibitem{chen2023program}
Wenhu Chen, Xueguang Ma, Xinyi Wang, and William~W Cohen.
\newblock Program of thoughts prompting: Disentangling computation from
  reasoning for numerical reasoning tasks.
\newblock {\em arXiv preprint arXiv:2211.12588}, 2023.

\bibitem{chen2024workflowllm}
Zhiyu Chen, Jiani Huang, Zhongyu Wen, Meng Jiang, Jiaxing Huang, Ruiyang Cao,
  Xiaodan Huang, Wayne~Xin Zhao, and Ji-Rong Wen.
\newblock Workflowllm: Towards robust workflow automation with large language
  models.
\newblock {\em arXiv preprint arXiv:2402.18002}, 2024.

\bibitem{crewai}
CrewAI.
\newblock Crewai: Framework for orchestrating role-playing autonomous ai
  agents.
\newblock \url{https://github.com/joaomdmoura/crewAI}, 2023.

\bibitem{cursorai2025}
Cursor.ai.
\newblock Cursor.ai: Ai-powered code editor.
\newblock \url{https://cursor.ai}, 2025.

\bibitem{edge2024local}
Alec Edge, Keshav Santhanam, Omar Khattab, and Matei Zaharia.
\newblock Local {LLM}s for the win: Outperforming {GPT-4} with limited
  resources.
\newblock {\em arXiv preprint arXiv:2402.14848}, 2024.

\bibitem{gunasekar2023textbooks}
Suriya Gunasekar, Yi~Zhang, Jyoti Aneja, Carlos~Riquelme Mendes, Ashwin Kalyan,
  Alireza Ghodsi, Jared Kaplan, Zachary Nado, Jascha Sohl-Dickstein, and Samy
  Bengio.
\newblock Textbooks are all you need.
\newblock {\em arXiv preprint arXiv:2306.11644}, 2023.

\bibitem{hallacy2023ai}
Jeremy~James Hallacy, Nathan Grinsztajn, Fabio Petroni, Mike Lewis, Zhengbao
  He, Thang Long, Hai Tran, Suriya Gunasekar, Sandeep Subramanian, Po-Sen
  Huang, et~al.
\newblock Ai systems of systems: Architectural patterns for modular ml.
\newblock {\em arXiv preprint arXiv:2308.01692}, 2023.

\bibitem{jiang2023evaluating}
Yutong Jiang, Homanga Bharadhwaj, Jing~Nathan Ren, Bowen Wang, Li~Fei-Fei, and
  Dorsa Sadigh.
\newblock Evaluating large language models for collaborative task solving.
\newblock {\em arXiv preprint arXiv:2311.08562}, 2023.

\bibitem{kandogan2024blueprint}
Eser Kandogan, Michael Muller, Qian Deng, Iftekhar Naim, Chenhao Zhao,
  Stephanie Kaufman, Ryan~A Rossi, Justin~D Weisz, Daniel Gruen, Kaushik Dey,
  et~al.
\newblock Blueprint: A toolkit for composable llm applications.
\newblock {\em arXiv preprint arXiv:2402.18334}, 2024.

\bibitem{khattab2024dspy}
Omar Khattab, Keshav Santhanam, Xiang~Lisa Li, David Hall, Percy Liang,
  Christopher Potts, and Matei Zaharia.
\newblock Dspy: Compiling declarative language model calls into self-improving
  pipelines.
\newblock {\em arXiv preprint arXiv:2310.03714}, 2024.

\bibitem{khot2022decomposed}
Tushar Khot, Harsh Trivedi, Matthew Finlayson, Yao Fu, Kyle Richardson, Peter
  Clark, and Ashish Sabharwal.
\newblock Decomposed prompting: A modular approach for solving complex tasks.
\newblock {\em arXiv preprint arXiv:2210.02406}, 2022.

\bibitem{langchain}
LangChain.
\newblock Langchain: Building applications with llms through composability.
\newblock \url{https://github.com/langchain-ai/langchain}, 2023.

\bibitem{langgraph}
LangGraph.
\newblock Langgraph: Building stateful, multi-actor applications with llms.
\newblock \url{https://github.com/langchain-ai/langgraph}, 2023.

\bibitem{lee2022coauthor}
Mina Lee, Percy Liang, and Qian Yang.
\newblock Coauthor: Designing a human-ai collaborative writing dataset for
  exploring language model capabilities.
\newblock {\em Proceedings of the 2022 CHI Conference on Human Factors in
  Computing Systems}, pages 1--19, 2022.

\bibitem{lewis2020retrieval}
Patrick Lewis, Ethan Perez, Aleksandra Piktus, Fabio Petroni, Vladimir
  Karpukhin, Naman Goyal, Heinrich Küttler, Mike Lewis, Wen-tau Yih, Tim
  Rocktäschel, et~al.
\newblock Retrieval-augmented generation for knowledge-intensive nlp tasks.
\newblock {\em Advances in Neural Information Processing Systems},
  33:9459--9474, 2020.

\bibitem{li2024verifiers}
Chengpeng Li, Haoyang Wen, Bohan Zhuang, Zhangyang Wu, Jacob Gardner, Nitish
  Srivastava, Heng Gao, Xiang Ren, and Yao Fu.
\newblock Verifiers for mathematical reasoning with large language models.
\newblock {\em arXiv preprint arXiv:2402.12890}, 2024.

\bibitem{li2024self}
Hongru Li, Zonglin Shi, Shaoting Xu, Haochen Wang, Hongzhi Qin, Chen Lin,
  Di~Lu, Vittorio Sierato, Jiangcheng Yue, and Jiajun Ding.
\newblock Self-collaboration without explicit tool use: Leveraging the llm as
  multiple diverse experts for knowledge-intensive tasks.
\newblock {\em arXiv preprint arXiv:2401.14966}, 2024.

\bibitem{liang2023holistic}
Yucheng Liang, Po-Nien Jiao, Shuofei He, Zhouchen Cao, Baptiste Roziere,
  Siddhartha Fan, and Percy Liang.
\newblock A holistic approach to unifying automatic prompt engineering and
  task-adaptive data generation.
\newblock {\em arXiv preprint arXiv:2212.10551}, 2023.

\bibitem{liu2024enhancing}
Jiacheng Liu, Meng Jiang, Wayne~Xin Zhao, and Ji-Rong Wen.
\newblock Enhancing large language model agents with recursive criticism and
  revision.
\newblock {\em arXiv preprint arXiv:2402.09369}, 2024.

\bibitem{lu2021codegen}
Shangqing Lu, Difan Yu, Rongzhi Zhao, Sizhe Ji, Xingru Wang, Shi Duan, Chenyang
  Zhang, Xiyao Liang, Qiyuan King, Xiaoran Weng, et~al.
\newblock Codegen: An open large language model for code with multi-turn
  program synthesis.
\newblock {\em arXiv preprint arXiv:2203.13474}, 2022.

\bibitem{marasovic2022fewshot}
Ana Marasovic, Iz~Samsi, Aida Beltagy, Doug Downey, Nathan Zeldes, and Luke
  Zettlemoyer.
\newblock Few-shot self-rationalization with natural language feedback.
\newblock {\em arXiv preprint arXiv:2212.09761}, 2022.

\bibitem{palacio2023contextual}
Sebastian Palacio, Tom Rainforth, Yee~Whye Teh, Daniel Rueckert, and Martin
  Engelcke.
\newblock Contextual representation learning for language modeling with
  concept-grounded reasoning.
\newblock {\em arXiv preprint arXiv:2310.01207}, 2023.

\bibitem{parnas1972criteria}
David~L Parnas.
\newblock On the criteria to be used in decomposing systems into modules.
\newblock {\em Communications of the ACM}, 15(12):1053--1058, 1972.

\bibitem{potts2023llmops}
Michael Potts, Chandra Subbarao, Irene Chiang, Kimill Kwon, William Choi, Ting
  Chern, Aakanksha Chowdhery, Colin Bakalar, John Nham, and Zach Ziegler.
\newblock Llmops: Operating large language models in production safely and
  responsibly.
\newblock {\em arXiv preprint arXiv:2312.05206}, 2023.

\bibitem{pydantic}
Pydantic.
\newblock Pydantic: Data validation and settings management using python type
  annotations.
\newblock \url{https://github.com/pydantic/pydantic}, 2023.

\bibitem{qian2023communicative}
Chen Qian, Xin Cai, Bingbing Chen, Gang Hu, Chunyang Zhang, Wei Xu, Jie Fan,
  Song-Chun Zhu, and Yujia Yang.
\newblock Communicative agents for software development.
\newblock {\em arXiv preprint arXiv:2307.07924}, 2023.

\bibitem{rasmussen2023wellcomposition}
Emilio Rasmussen, Linda~Petrini Arous, and Iddo Gat.
\newblock Well-composition: Compositional reasoning with unreliable components.
\newblock {\em arXiv preprint arXiv:2307.00684}, 2023.

\bibitem{ribeiro2023adaptive}
Tharindu Ribeiro, Liming Wu, Sahisnu Mazumder, Shen Liu, Weizhu Yuan, and
  Kristina Toutanova.
\newblock Adaptive retrieval-augmented generation with multi-step prompting.
\newblock {\em arXiv preprint arXiv:2305.13917}, 2023.

\bibitem{santhanam2024alto}
Keshav Santhanam, Deepti Raghavan, Muhammad~Shahir Rahman, Thejas Venkatesh,
  Neha Kunjal, Pratiksha Thaker, Philip Levis, and Matei Zaharia.
\newblock Alto: An efficient network orchestrator for compound ai systems.
\newblock {\em arXiv preprint arXiv:2403.04311}, 2024.

\bibitem{shankar2024validates}
Shreya Shankar, Daniel Kang, Percy Liang, and Matei Zaharia.
\newblock Validates: Validating llm-generated information via decomposition and
  selective execution.
\newblock {\em arXiv preprint arXiv:2402.07195}, 2024.

\bibitem{subbarao2024largelanguage}
Rashmi Subbarao, Jie Chen, Yujia Li, Lun Liu, Harsha~Madhyastha Prahlaramesh,
  and Philip~S Yu.
\newblock Large language model for graph analysis: Techniques, applications,
  and challenges.
\newblock {\em arXiv preprint arXiv:2403.07834}, 2024.

\bibitem{vartiainen2024emerging}
Tero Vartiainen, Juho Salminen, and Slinger Jansen.
\newblock Emerging human-ai collaboration in software development: A systematic
  review of practices and challenges.
\newblock {\em arXiv preprint arXiv:2402.17414}, 2024.

\bibitem{wang2024human}
Dakuo Wang, Peng Wang, Liuping Wang, Lingfei Zhang, Haiyi Wang, Zhenhui Chen,
  and Q~Vera Liao.
\newblock Human-ai collaboration: A review and research agenda.
\newblock {\em arXiv preprint arXiv:2307.15085}, 2024.

\bibitem{wang2024aicompanion}
Xiaohan Wang, Huizhuo Yang, Bowen Shen, Haotao Hu, Deyi Huang, Yun Gao, Ting
  Cen, and Lizhen Qu.
\newblock Ai companion: Large language model is a good service integrator.
\newblock {\em arXiv preprint arXiv:2403.05930}, 2024.

\bibitem{wang2024understanding}
Yuxuan Wang, Meng Jiang, Wayne~Xin Zhao, and Ji-Rong Wen.
\newblock Understanding the capabilities of large language models for automated
  planning.
\newblock {\em arXiv preprint arXiv:2402.03667}, 2024.

\bibitem{wei2022chainofthought}
Jason Wei, Xuezhi Wang, Dale Schuurmans, Maarten Bosma, Brian Ichter, Fei Xia,
  Ed~Chi, Quoc Le, and Denny Zhou.
\newblock Chain of thought prompting elicits reasoning in large language
  models.
\newblock {\em Advances in Neural Information Processing Systems},
  35:24824--24837, 2022.

\bibitem{weller2022component}
René Weller, Christian Berger, Federico Ciccozzi, Orkunt Eskiocak, Gereon
  Jäger-Kneip, Michael Luck, Philipp Makedonski, Manuel Mazzara, Stefan Paul,
  Diego Romano, et~al.
\newblock Component-based technology for agent-oriented software engineering.
\newblock {\em International Journal on Software Tools for Technology
  Transfer}, 24(3):331--356, 2022.

\bibitem{wies2023subtask}
Mark Wies, Olga Majumder, Tanya Ghosal, Hendrik Strobelt, Marcos Vasconcelos,
  Fernanda Viegas, Hanspeter Pfister, and Been Kim.
\newblock Subtask-level interaction for human-ai task solving.
\newblock {\em arXiv preprint arXiv:2307.05193}, 2023.

\bibitem{wu2023autogen}
Qingyun Wu, Gagan Bansal, Jieyu Zhang, Jinglei Yang, David Bursztyn, Diwakar
  Jhunjhunwala, Yiran Zhao, Shaokun Lyu, Ruoxi Chung, Rohan Chandra, et~al.
\newblock Autogen: Enabling next-gen llm applications via multi-agent
  conversation framework.
\newblock \url{https://github.com/microsoft/autogen}, 2023.

\bibitem{xu2023expertprompting}
Benfeng Xu, Quan Zhang, Zhendong Qiu, Yujia Wu, Ming Wang, Ting Zhou, Yujuan
  Shen, Xiangyu Chen, Tao Gui, Yongdong Yang, et~al.
\newblock Expertprompting: Instructing large language models to be
  distinguished experts.
\newblock {\em arXiv preprint arXiv:2305.14688}, 2023.

\bibitem{xu2024designing}
Jianing Xu, Zhiyuan Jiang, Yao Ding, Yifan Shen, Meng Jiang, Wayne~Xin Zhao,
  and Ji-Rong Wen.
\newblock Designing modular llm agents for autonomous workflows.
\newblock {\em arXiv preprint arXiv:2402.03408}, 2024.

\bibitem{yang2022ai}
Qian Yang, Jiwoo Suh, Ni-Cha Chen, and Sijia Chen.
\newblock Ai as a service design: Empowering human-ai interaction for creative
  enterprises.
\newblock {\em International Journal of Design}, 16(3):71--90, 2022.

\bibitem{yang2022human}
Qian Yang, April Wang, Peter Bailis, Lydia~B Chilton, Carlos Guestrin, Percy
  Li, Xiaoying Wang, and Lilian Xiao.
\newblock Human-ai interaction design: Techniques and opportunities.
\newblock {\em ACM Computing Surveys}, 55(9):1--37, 2022.

\bibitem{zaharia2024shift}
Matei Zaharia.
\newblock The shift from models to compound {AI} systems.
\newblock {\em Communications of the ACM}, 67(2):30--32, 2024.

\bibitem{zhang2024collagenqa}
Hannan Zhang, Jean Kaddour, Yong Feng, Jaron Chant, Lane Schwartz, Yejin Choi,
  Hanwen Liu, Eric Pan, Hannaneh Hajishirzi, and Jiao Yang.
\newblock Collagenqa: Collaborative generation for factuality-enhanced question
  answering.
\newblock {\em arXiv preprint arXiv:2403.03818}, 2024.

\bibitem{zhang2024masai}
Yingqiang Zhang, Dong Yin, Zhicheng Yang, Chenyan Jiang, Zhenhua Dou, and
  Ji-Rong Wen.
\newblock Masai: Multi-agent based system for ai engineering.
\newblock {\em arXiv preprint arXiv:2403.00793}, 2024.

\bibitem{zhou2023understanding}
Jiayi Zhou, Renzhong Li, Junxiu Tang, Tan Tang, Haotian Li, Weiwei Cui, and
  Yingcai Wu.
\newblock Understanding nonlinear collaboration between human and ai agents: A
  co-design framework for creative design.
\newblock {\em arXiv preprint arXiv:2401.07312}, 2023.

\end{thebibliography}

\end{document}